\documentclass[10pt,twocolumn,letterpaper]{article}

\usepackage[pagenumbers]{cvpr} 

\usepackage{lineno}
\usepackage{marvosym}
\makeatletter
\robustify\@latex@warning@no@line
\makeatother
\usepackage{authblk}
\usepackage[colorlinks,linkcolor=red]{hyperref}
\usepackage{multirow}
\def\paperID{4919} 
\def\confName{CVPR}
\def\confYear{2025}

\title{SpatialDreamer: Self-supervised Stereo Video Synthesis from Monocular Input}

\author{
Zhen Lv\textsuperscript{1}, Yangqi Long\textsuperscript{1}, Congzhentao Huang\textsuperscript{1}, Cao Li\textsuperscript{1}, Chengfei Lv\textsuperscript{1}\textsuperscript{\dag}, Hao Ren\textsuperscript{1}, Dian Zheng\textsuperscript{2}\vspace{-3mm}\\
\textsuperscript{1}Alibaba Group; \textsuperscript{2}Sun Yat-sen University}

\begin{document}
\maketitle
\begin{abstract}
Stereo video synthesis from monocular input is challenging in spatial computing and virtual reality due to the lack of high-quality stereo video pairs for training and the difficulty of maintaining spatio-temporal consistency between frames.
Existing methods primarily address these issues by directly applying novel view synthesis (NVS) techniques to video, while facing limitations such as the inability to effectively represent dynamic scenes and the requirement for extensive training data.
In this paper, we introduce a novel self-supervised stereo video synthesis paradigm via a video diffusion model, termed \textbf{SpatialDreamer}, which meets the challenges head-on. Firstly, to address the stereo video data insufficiency, we propose a \textbf{D}epth based \textbf{V}ideo \textbf{G}eneration module \textbf{DVG}, which employs a forward-backward rendering mechanism to generate paired videos with geometric and temporal priors. Leveraging data generated by DVG, we propose RefinerNet along with a self-supervised synthetic framework designed to facilitate efficient and dedicated training.
More importantly, we devise a consistency control module, which consists of a metric of stereo deviation strength and a \textbf{T}emporal \textbf{I}nteraction \textbf{L}earning module \textbf{TIL} for geometric and temporal consistency ensurance respectively. We evaluated the proposed method against various benchmark methods, with the results showcasing its superior performance. 
Our project website is at: \url{https://spatialdreamer.github.io}.
\end{abstract}  
\let\thefootnote\relax\footnotetext{\textsuperscript{\dag} Corresponding author.}
\begin{NoHyper}
\section{Introduction}
\label{sec:intro}
Stereo video synthesis from a monocular input aims to generate the target-view video based on the given view with geometric and spatio–temporal consistency, which has a wide range of applications in 3D model reconstruction~\cite{areview}, 3D movie production~\cite{Low-cost360, gao2021dynamicviewsynthesisdynamic}, and Apple Vision Pro (AVP)~\cite{avp} like virtual reality content.

\begin{figure}[t]
\centering
\includegraphics[width=0.99\linewidth]{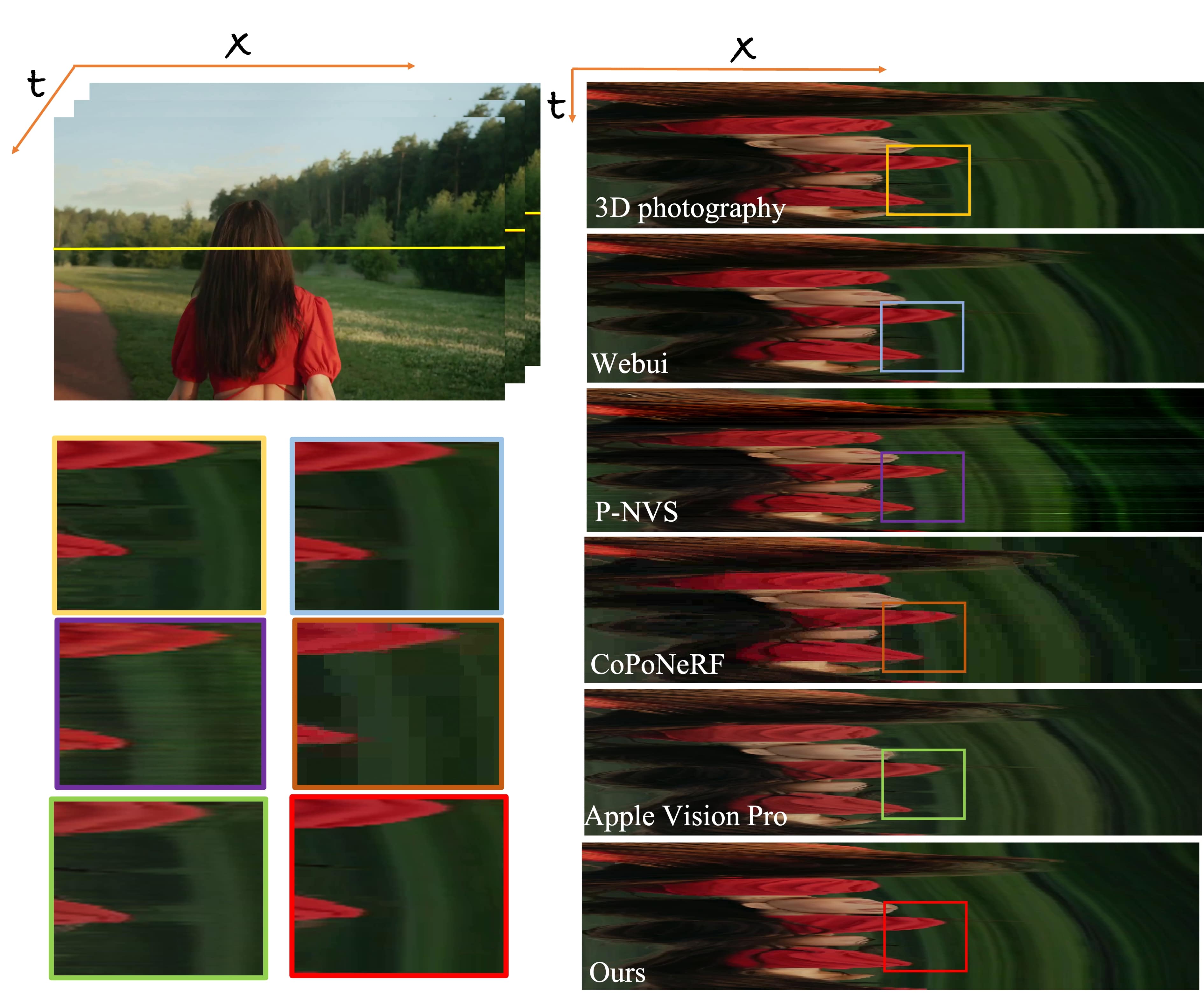}
\caption{Visualization of the temporal consistency in stereo video generated by different methods. We extract the yellow line of each frame and stack them together. A good result should show a natural transition in the $t$ dimension.}
\label{fig:cover}
\vspace{-2mm}
\end{figure}


The primary difficulties of this task stem from the lack of adequate high-quality paired stereo videos for training, and the challenge of preserving the geometric consistency between two views as well as maintaining the temporal consistency across generated frames.
Traditional methods for generating stereo content involve capturing scenes using dual-camera setups~\cite{ViewSynthesis,imageDistortions,StereoscopicVideoSynthesis,DIBR}. However, the acquisition of these images involves the use of a professional-grade camera, resulting in substantial costs. 
Recent advancements in deep learning mainly handle these problems by directly applying single to stereo or multi-view image generation techniques to video, while often facing challenges related to geometric and spatio-temporal inconsistencies~\cite{rombach2022high, wang2023learning, han2022single, wang2024videocomposer,mildenhall2020nerfrepresentingscenesneural, kerbl20233dgaussiansplattingrealtime}. As shown in Fig~\ref{fig:cover}, 
methods based on NVS are challenging to maintain accurate spatio-temporal consistency, and even the recently released AVP 3D photo converter~\cite{wwdc} may inevitably result in content flickering and inconsistency when applied to video synthesis.
This is due to the difficulty of handling the temporal coherence between frames and the spatial information between paired views in complex dynamic scenes.

In this paper, we propose a self-supervised stereo video synthesis paradigm via a video diffusion model, termed SpatialDreamer, which will meet the data insufficiency and spatio-temporal inconsistency at once. 
Firstly, we design a \textbf{D}epth based \textbf{V}ideo data \textbf{G}eneration module \textbf{DVG} to handle the problem of data insufficiency. 
Similar to \cite{han2022single,wang2023learning} , DVG constructs a training pair of two views using forward-backward rendering without the need for data annotation ~\cite{DIBR,imagelabelling, imageDistortions}, while \cite{han2022single,wang2023learning}  only render individual images, resulting in flickering effects in generated video~\cite{LargeOcclusionStereo}.
In this paper, we improve the generation of video data for refining stereo occlusion masks by leveraging the inter-frame motion obtained from optical flow~\cite{jeong2024ocaiimprovingopticalflow}.
This methodology enables DVG to produce paired video data that maintains both geometric and temporal consistency.
Leveraging paired video generated by DVG, we propose RefinerNet, along with a self-supervised video synthesis framework, to enable efficient and targeted training. More importantly, with access to sufficient paired videos, 
we devise a consistency control module, which consists of a metric of stereo deviation strength and a \textbf{T}emporal \textbf{I}nteraction \textbf{L}earning module \textbf{TIL}. 
The stereo deviation strength aims to enable the generation of stereo videos in diverse real-world scenarios, with a stereo-aware loss further supervising the model to learn the magnitude of difference between paired view features in the latent space.
TIL integrates the latent features from long-temporal frames as a global information to augment the temporal coherence of the generated results.

Extensive experiments compared with various benchmark methods show that we achieve state-of-the-art  performance. Notably, our SpatialDreamer meets the demand of real-world application without jitter, geometric and temporal inconsistency, which even beats AVP 3D converter and any open-source stereo video synthesis methods.
We summarize our contributions as follows:
\begin{itemize}
\item A novel self-supervised stereo video synthesis framework, SpatialDreamer, is proposed, which is robust across a wide range of scenes and dynamic content in video. Moreover, equipping the DVG module, we build a stereo video dataset using a self-supervised approach.
\item A consistency control module is devised, which consists of a metric of stereo deviation strength and a temporal interaction learning module, ensuring geometric and temporal consistency in video sequences.
\item The sufficient experiments and quantitative results demonstrate that the proposed method outperforms the ate-of-the-art methods, even beats AVP.
\end{itemize}

\section{Related Work}
\begin{figure*}[t]
\centering
\includegraphics[width=0.95\textwidth]{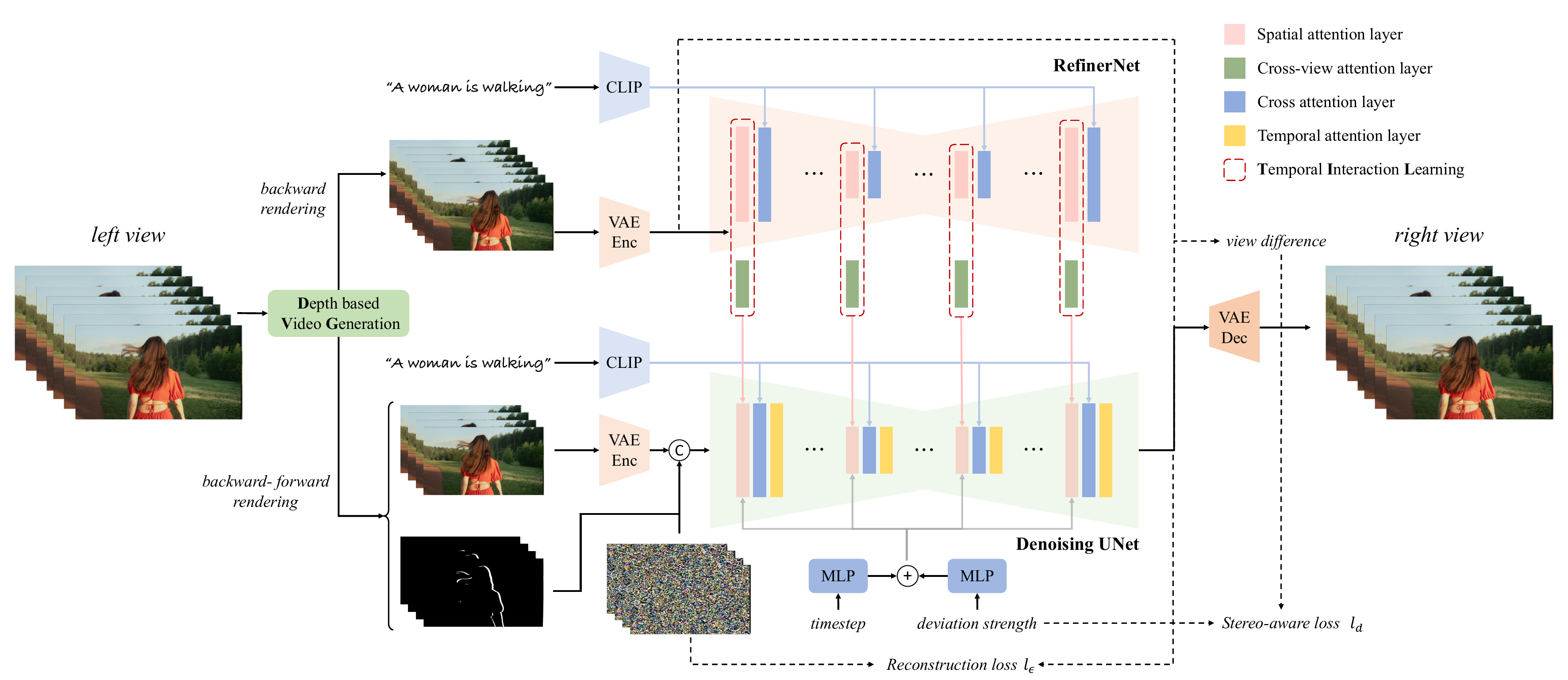}
\caption{Overview of the proposed method, given left view video, the target view video is rendered, encoded, and concatenated with multi-frame noise, followed by the denoising U-Net architecture (i.e., SVD). 
The reference view images are fed into RefinerNet, through which the spatial features are extracted. (\textbf{Notably, the left and right views are the target and reference view for both training and inference}). The temporal interaction learning module integrates the latent features from long-temporal frames, and the deviation strength is projected as positional embedding and added to the time step embedding.  
Finally, the variational autoencoder decoder decodes the result into a video clip.}
\vspace{-4mm}
\label{fig_flowmap}
\end{figure*}
\subsection{Single-image Novel View Synthesis}
Early studies address the single-image novel view synthesis task without relying on explicit 3D representation, such as End-to-end Synthesis~\cite{wiles2020synsin}, Infinite-Nature~\cite{liu2021infinite} and Animatable 3D Characters~\cite{weng2019photo}. 
These methods lack explicit 3D representations for scenes, leading to inconsistent views and causing flickering and blurring, especially in occluded regions.
Layer-based methods, such as 3D-photography~\cite{shih20203d} and SLIDE~\cite{jampani2021slide} represent a 3D scene using discrete layers. Yet, the synthesized views may lack accuracy in texture and geometric consistency across different perspectives. SMPI~\cite{tucker2020single}, AdaMPI~\cite{han2022single} and SinMPI~\cite{pu2023sinmpinovelviewsynthesis} represent the 3D scene using multiple planes, while their rendered results may suffer from depth discretization artifacts and repetitive texture patterns. 
To reduce the reliance on large labeled datasets, Shi et.al~\cite{shi2021selfsupervised},
Wang et al.~\cite{wang2023learning} and NVSVDE-Net~\cite{bello2024novel} explore self-supervised learning for creating novel view images from single image.
Although they produce photorealistic results, they struggle with temporal and spatial consistency, especially in dynamic scenes.

As for our approach, we design a depth-based video generation module that generates paired videos with geometric and temporal priors. Additionally, Our method ensures multi-view consistency for the entire scene by employing a consistency control module.
\subsection{Multi-view Novel View Synthesis}

Early approaches in this field frequently utilized geometric methods, such as multi-view stereo reconstruction~\cite{1640800, jin2005multi} and structure-from-motion techniques~\cite{schonberger2016structure, ozyecsil2017survey}, to generate novel viewpoints from multiple images captured at different angles. 
NeRF-based methods were proposed to generate novel views based on fewer input images, including SinNeRF~\cite{xu2022sinnerftrainingneuralradiance}, CoPoNeRF~\cite{hong2024unifying} and NViST~\cite{jang2024nvistwildnewview} .
3D Gaussian Splatting~\cite{kerbl20233dgaussiansplattingrealtime} presented an explicit representation of the scene. Gaussian-based methods, such as pixelSplat~\cite{charatan23pixelsplat} and MVSplat~\cite{chen2024mvsplat}, can produce high-quality synthesis results from multi-view images.
Recently, researchers have begun exploring the application of advanced diffusion models for novel view synthesis, driven by their remarkable success in this area. Notable approaches such as Photoconsistent-NVS~\cite{yu2023longterm}, MultiDiff~\cite{muller2024multidiff}, GCD~\cite{vanhoorick2024gcd} and NVS-Solver~\cite{you2024nvs} generate image sequences based on predefined camera trajectories using either single or multiple input images. 
While these methods have shown promising results, they often struggle with dependence on dense scene geometry, handling complex scene dynamics, and generalization challenges~\cite{kerbl20233dgaussiansplattingrealtime}. 
Moreover, the challenge of optimizing NeRF/GS with sparse views hampers the ability of multi-view NVS to produce high-quality results.

In contrast, we explore a video diffusion model to achieve stereo video synthesis, while preserving geometric structures through a RefinerNet based framework.

\subsection{Video Synthesis via Diffusion Models}

For video synthesis, 
previous works mainly expand image models to video, exploiting the image generation capability in the pretrained image diffusion models~\cite{liu2024sorareviewbackgroundtechnology}. The video diffusion model (VDM)~\cite{ho2022video} and Video LDM~\cite{blattmann2023align} are proposed to adopt a 3D U-Net structure for joint image and video training.
Subsequent works~\cite{singer2022make,guo2023animatediff,guo2023animatediff,hu2023animate,dai2023animateanything} also follow this insight with temporal attention, and \cite{wu2023tune,wang2023zero,khachatryan2023text2video,qi2023fatezero} further apply cross-frame attention to improve the consistency of the generated video.
Another approach to promote temporal consistency is encoding video into 2D images~\cite{bar2022text2live, ouyang2023codef}, which has been mainly explored in the video editing area. 
Although these models produce consistent results, they often require per-video optimization, which takes a long time, and would result in degradation when dealing with large motion.
Most recently, \cite{kondratyuk2023videopoet,gupta2023photorealistic,videoworldsimulators2024} employ a transformer~\cite{vaswani2017attention} architecture to model the temporal and spatial relations, demonstrating excellent video generation capabilities. However, these models need very large scale video datasets for training, which can be difficult to collect.

Beyond temporal information modeling, some works~\cite{esser2023structure,wang2024videocomposer,chen2023control,zhao2023controlvideo} have introduced aligned spatial guidance to enhance the stability of the generated videos. The proposed framework also uses spatial guidance. Differing from those methods which require spatially aligned controls, a spatially unaligned reference image is used via a cross-attention mechanism. Furthermore, a metric of stereo deviation strength is introduced to strengthen the spatial constraint, resulting in more consistent results.

\section{SpatialDreamer}
Given an image sequence and stereoscopic pose describing the scene depth information, the proposed method generates a stereo pair. In this section, we first provide a brief introduction about SVD~\cite{blattmann2023stable}. Secondly, we offer detailed information of our video generation module DVG, which can generate sufficient paired video with geometric and temporal consistency. Using DVG, we present the framework for self-supervised learning-based video synthesis to support efficient and dedicated training. More importantly, to ensure consistency between frames, we design a consistency control module for geometric and temporal consistency ensurance. Finally, we present the detailed training process.

\subsection{Preliminaries}
The challenge in video generation lies in accurately modeling the spatial-temporal dynamics, involving spatial relationships within frames and temporal relationships across frames.
To address this challenge, Stable Video Diffusion (SVD)~\cite{blattmann2023stable} introduces temporal convolution and attention layers. SVD follows the latent diffusion model (LDM) to encode video pixels into the latent space, enabling a more efficient denoising process. Given a video $I_v$, encoder $\mathcal{E}$ encodes $I_v$ into the latent space as $z_0 = \mathcal{E}(I_v)$. Gaussian noise $\epsilon$ is then added in the Markov process:
\begin{equation} 
z_t=\sqrt{\bar{\alpha}_t}z_0+\sqrt{1-\bar{\alpha}_t}\epsilon_t
\end{equation} 
where $\bar{\alpha}_0,\ldots,\bar{\alpha}_T$ are the pre-defined noise schedules for T steps in the Markov process. In particular, SVD is trained by the $v$-prediction formulation with the mean squared error (MSE) between the ground truth and its prediction in the latent space, known as $l_{\epsilon}$. 
\subsection{Depth based Video Generation}
To create training pairs of images along with their corresponding occluded masks, the following self-supervised mechanism is employed utilizing real images. Firstly, the depth of the given video sample is estimated~\cite{Ranftl2022}. Secondly, the reference view image $x_1$ is rendered into a masked one under the target viewpoint $P_2$, and then the trained inpainting model~\cite{stacchio2023stableinpainting} is used to fill these occluded regions to obtain the novel viewpoint  $x_2$~\cite{niklaus2020softmaxsplattingvideoframe}. Finally, $x_2$ is rendered to the original viewpoint $P_1$ via a forward rendering. After this, we can obtain the masked image $\widetilde{x}_1$ with mask $M$ under viewpoint $P_1$, as well as the inpainted image $x_2$ under viewpoint $P_2$. The mask $M$ and masked image $\widetilde{x}_1$ are considered to be the conditions, with the image $x_1$ serving as the ground truth for the diffusion model. This approach involves encoding the inpainted image $x_2$ into latent feature representations and feeding it into RefinerNet.

However, the aforementioned approach encounters challenges in addressing inconsistencies among generated data frames, resulting in observable artifacts such as jittering and flickering. 
To tackle this issue, we leverage optical flow~\cite{raft} to capture motion between successive video frames.
Specifically, by leveraging the optical flow between the reference image $x_1^{t}$ at time $t$ and its adjacent frames, we establish the pixel correspondences between the image $x_1^{t}$ and the images $x_1^{t-1}$ and $x_1^{t+1}$, respectively. Particularly, these optical flows enable computing a confidence map via forward-backward consistency~\cite{1023808}, which allows us to more accurately merge the stereo occlusions in the masks between adjacent frames~\cite{occ_refine}. 
Subsequently, the refined occlusion mask $m^{t}$ is generated by sampling from the occlusion maps $m^{t-1}$ and $m^{t+1}$, leveraging the pixel correspondences between consecutive frames under the reference view, as described by the following equation.

\begin{equation}
m^t(i, j) =
\begin{cases}
1, & \begin{aligned}[t]
\sum\limits_{k \in \{t-1, t+1\}} &m^k(i+u, j+v) \cdot  \\
& C(i, j) \geq 1
\end{aligned} \\
m^t(i, j), & \quad else
\end{cases}
\end{equation}
where $u$ and $v$ are the x and y components of the optical flow vector, and $C$ represents the optical flow confidence measure.
By propagating information from occluded areas across frames, this inter-frame continuity facilitates a more precise refinement of occluded regions in paired view videos and enforces temporal smoothness and coherence within the self-supervised framework, effectively mitigating the undesirable effects of jitter and flicker.

As shown in Figure~\ref{fig_train_infer}, the difference between the training and inference procedures is that the former uses both backward and forward rendering whereas the latter uses only forward rendering to obtain an image under the target view. 
\begin{figure}[t]
\centering
\includegraphics[width=0.95\columnwidth]{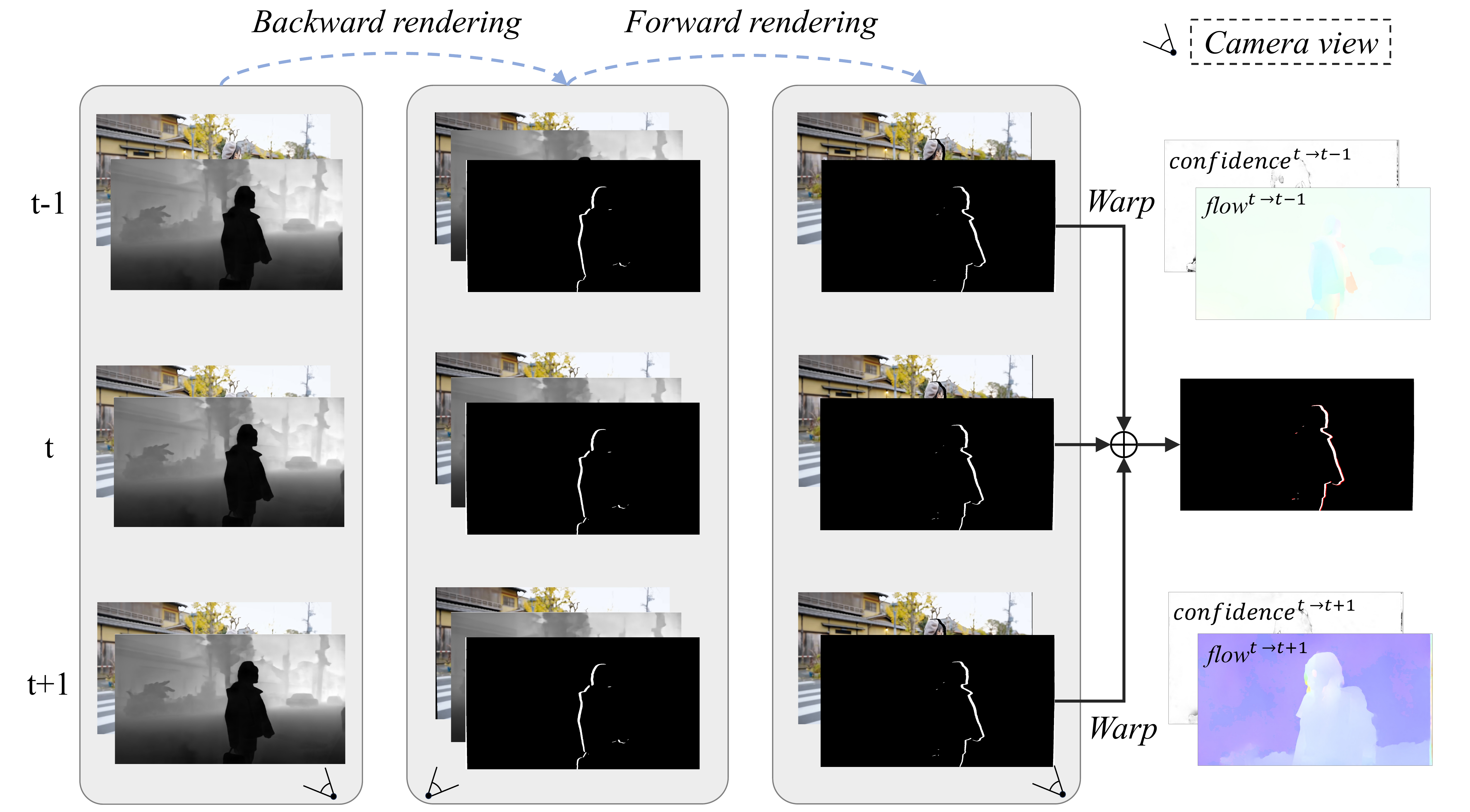} 
\caption{DVG. The temporal motion can be utilized to refine the occluded regions in the current frame, thereby providing smoother images and more consistent occlusion in the temporal sequence.
}
\label{fig_train_infer}
\end{figure}

\subsection{Self-Supervised Stereo Video Synthesis framework}
Leveraging the generated paired view videos, we propose a self-supervised framework for stereo video synthesis, as illustrated in Figure~\ref{fig_flowmap}. This framework commences with the rendering of the target view video that incorporates occlusions, followed by the application of a spatio-temporal method to effectively inpaint the missing regions, ultimately yielding a complete non-occluded target view video.


To learn the distribution of the feature space between the paired view images, we devise RefinerNet, which is modeled after the architecture of the denoising U-Net, excluding the temporal layer.  
RefinerNet adopts the weight initialization from the original SD2.1 model~\cite{rombach2022high}. As depicted in Figure 2, a spatial attention layer is employed instead of the self-attention layer. Given a feature map $z_{t}\in \mathbb{R}^{t\times h\times w\times c}$ from the denoising U-Net and another feature map $z_{r}\in \mathbb{R}^{t\times h\times w\times c}$ from RefinerNet, they are concatenated along the h dimension, and then self-attention is applied. Subsequently, the first half of the feature map is retrieved as the output. One great advantage of this approach is that it enables RefinerNet to maintain the feature representation capability inherent in the original SD Unet, extracting reference view image information. In addition, the denoising U-Net can adaptively learn features from RefinerNet that are correlated in the same feature space, owing to the fundamentally similar network architectures between the denoising U-Net and RefinerNet. This not only ensures a robust initialization for the feature representation but also enhances the learning ability of the entire network.



\subsection{Consistency Control Module}
\subsubsection{Temporal Interaction Learning module}
While our RefinerNet-based approach providing spatial guidance, paired view features are still merged independently, posing challenges for temporal consistency and potentially causing texture discrepancies or odd results between adjacent frames.

Inspired by previous studies~\cite{cao2023masactrltuning,hertz2022imageeditingcross}, we propose an attention-based temporal interaction learning module that learns the distribution of the feature space from adjacent frames, named TIL.
Specifically, given latent features of $N_r$ adjacent images under reference view $z_i^t$, where $i$ = 1, 2, ..., $N_r$ and the latent feature of the reference view image $z_r^t$ at the time step $t$, we augment the $z_r^t$ to $aug_r^t$ as:
\begin{equation} 
aug^t_r = \lambda \cdot \mathrm{Attn}_{r,r}  + (1-\lambda) \cdot  {\frac{1}{N_r}} \sum_{i = 1}^{N_r}\mathrm{Attn}_{r,i}
\label{atten}
\end{equation}
where $\lambda \in [0, 1]$ and $\mathrm{Attn}_{i,j}$ defines attention between two latent features $z_i$ and $z_j$. 
This module blends self-attention of $z_r^t$ in RefinerNet's U-Net block with the cross-view attention between $z_r^t$ and each adjacent view $z_i^t$.
The cross-view attention aligns the feature of all adjacent frames to the reference view 
while the self-attention helps each reference image retain distinctiveness. 
The augmented reference feature $aug_r^t$ is then fed into spatial attention layer to assist U-Net network learning.

\subsubsection{Stereo Deviation Strength}
During the training, we noticed that different viewpoints impact the 3D impression of the scenes in the generated videos. However, relying solely on a given viewpoint as a guide for generating stereoscopic videos is insufficient. This is because videos with the same viewpoint may exhibit diverse depth ranges and variations, depending on the scene~\cite{imageDistortions}. Therefore, pose alone cannot accurately control the 3D effect of a scene. Consequently, we introduce a metric called the stereo deviation strength, which quantitatively assesses the binocular disparity in a scene and facilitates the creation of controllable stereo vision:
\begin{equation} 
{s}(z) = \left|{z}_0 -{z}_{ref}\right|
\end{equation}
where the stereo deviation strength quantifies the latent differences between reference view  $z_{ref}$ and target view  $z_{0}$. Similar to the time step, the deviation strength is projected into a positional embedding and added to each frame in the residual block to ensure uniform application of the deviation strength to every frame. The impact of stereo deviation strength guidance on the results is illustrated in Figure~\ref{fig_devia_exp}.

Moreover, for better convergence, a stereo-aware loss function is proposed to directly supervise the disparity difference:
\begin{equation} 
{l}_d = || s({z}_0) -{ s(\hat{z}}_{0})||^2_2 
\end{equation}

where $\hat{z}_{0}$ represents the estimated clean video latent
$z_0$, which can be obtained by:
\begin{equation} 
{\hat{z}_{0}} = { {z}_{0} } - \frac{ \sqrt{1- \overline{\alpha}_t} \mathbf{\epsilon}_{\theta}(z_t,c,t)}{\sqrt{\overline{\alpha}_t}}
\end{equation} 
where $\overline{\alpha}_t$ = $\prod_{i=1}^{t}(1-\beta_{t})$, and the noise prediction loss is combined with the
stereo-aware loss by a scaling factor:
\begin{equation} 
l = l_{\epsilon} + \lambda \cdot l_d
\end{equation} 

\begin{figure}[t]
\centering
\includegraphics[width=0.9\columnwidth]{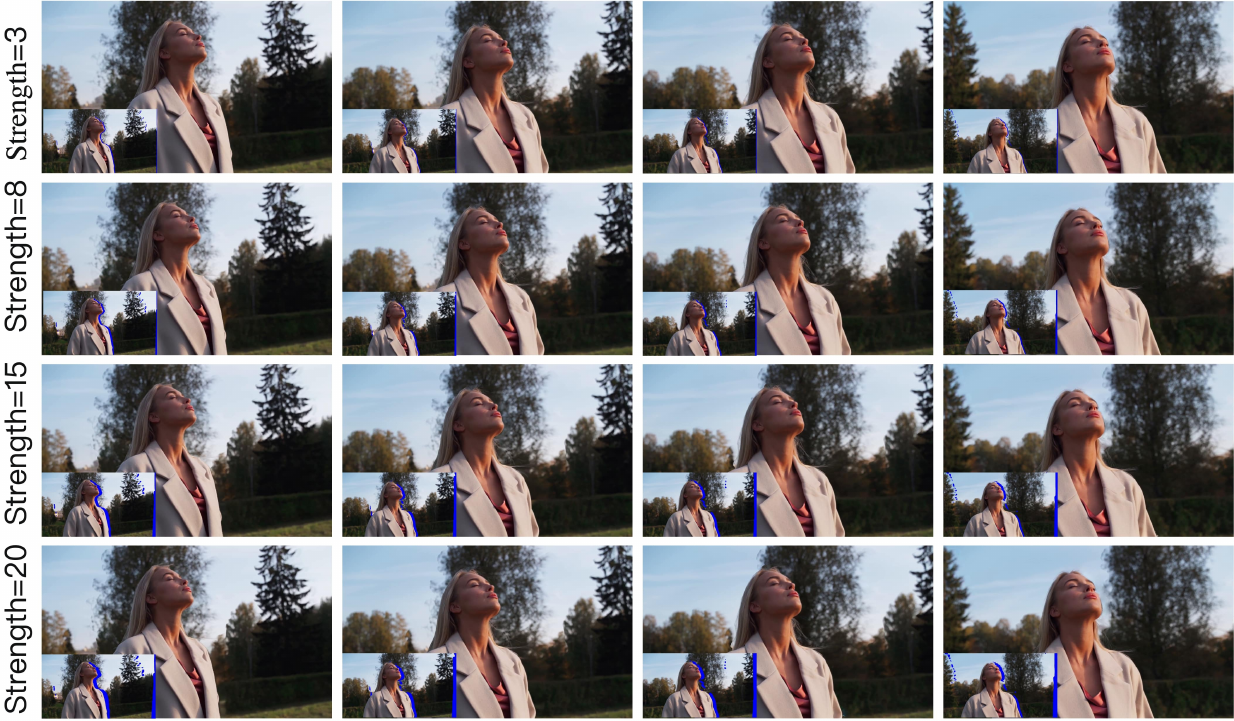} 
\caption{Stereo deviation strength guidance examples. Augmenting the deviation strength enhances the 3D photography effect, but excessive deviation strength may lead to image distortions or even lower the quality of stereoscopic image.}
\label{fig_devia_exp}
\end{figure}

\subsection{Training Strategy}
The training process is divided into two stages. In the first training stage, we focus on individual frames from videos. The temporal layer in denoising U-Net is frozen, and cross-view attention in TIL module is removed. The RefinerNet model and the denoising U-Net are initialized with pretrained weights from SD2.1~\cite{rombach2022high} and SVD~\cite{blattmann2023stable} respectively. The weights of the variational autoencoder encoder and decoder, as well as the contrastive language-image pre-training (CLIP) image encoder, are all kept fixed. The aim of this stage is for the model to learn to synthesize new viewpoint images under the condition of a reference image and a new viewpoint pose. In the second stage, we train the temporal layer and cross-view attention in TIL module with video sequences. This enables the model to capture temporal context information efficiently. The input for the model is the 8-frame video clip and $N_r$ for TIL is set to 8. 

\section{Experiments}

\subsection{Datasets, Baselines and Metrics}
We conduct our experiments on image and video levels with RealEstate10K~\cite{zhou2018} and self-collected video data. RealEstate10K is a large, widely used dataset with camera poses, corresponding to 10 million frames derived from about 80,000 video clips, gathered from about 10,000 YouTube videos. As there is not open-source stereo video benchmark, we collect 1500 monocular videos with 1400 for training, 100 for testing. 
For image synthesis, we compared the proposed method with recent open-source SOTA methods: 3D-photography~\cite{shih20203d}, SynSin~\cite{wiles2020synsin}, 
AdaMPI~\cite{han2022single},
SinMPI~\cite{pu2023sinmpinovelviewsynthesis},
Wang et al.~\cite{wang2023learning}, Photoconsistent-NVS(P-NVS)~\cite{yu2023longterm}, NVSVDE-Net~\cite{bello2024novel}, CoPoNeRF~\cite{hong2024unifying} and MVSplat~\cite{chen2024mvsplat}.
For a fair comparison, the 3D photography method, AdaMPI, SinMPI, and the proposed method all used MiDaS~\cite{Ranftl2022} for depth estimation. 
Following the method of  3D-photography~\cite{shih20203d}, we measured the accuracy of the generated target views with regard to the ground-truth images using the three metrics of the learned perceptual image patch similarity (LPIPS), peak signal-to-noise ratio (PSNR), and structural similarity index measure (SSIM).


For the video synthesis, we picked several representative approaches, i.e., the method of  3D-photography~\cite{shih20203d}, the open-source code repository webui-depthmap\footnote{https://github.com/thygate/stable-diffusion-webui-depthmap-script
}, P-NVS~\cite{yu2023longterm}, CoPoNeRF~\cite{hong2024unifying}, NVS-Solver~\cite{you2024nvs} and AVP~\cite{avp}. Moreover, we conducted the same depth estimation method~\cite{Ranftl2022} for a fair comparison of  3D-photography~\cite{shih20203d}, webui-depthmap and NVS-Solver~\cite{you2024nvs}. 
We employ the 
FVD~\cite{DBLP:conf/iclr/UnterthinerSKMM19} score to measure the perceptual similarity between input videos and outputs and report the flow warping error $E_{warp}$~\cite{lai2018learning} to assess the temporal consistency of the resulting video sequences.

\subsection{Implementation}
We employed the AdamW~\cite{loshchilov2017decoupled}  optimizer with a learning rate of $1\times{10}^{-5}$ to train the model on two NVIDIA A800 GPUs. 
In the first training stage, individual video frames were resized and center-cropped to a resolution of 1024 × 1024.
We trained the model for 100,000 iterations with a batch size of 8. In the second training stage, the videos were first split according to the transitions to ensure that a scene only appeared in one video clip. We then formed each video clip for training with eight frames. Finally, the temporal layer and TIL module was trained for 10,000 steps with a batch size of 2. The stereo-aware loss scaling factor $\lambda$ was set to 0.001 and $\lambda$ in Eq.~\ref{atten} was set as 0.6.

\subsection{Qualitative Results}
\subsubsection{Stereo Image Synthesis} We visually compared different methods for generating new view images from a target viewpoint.
We qualitatively compared the newly synthesized images in Figure~\ref{fig_compare_img}.  3D-photography fails to resolve the image distortion problem, and the image appears to be enhanced only slightly by the rendering. SynSin generates the most blurry details and loses a lot of information, compared with the input view. Neither AdaMPI nor SinMPI  can generate fine enough details, and the results contain noticeable artifacts. CoPoNeRF tends to produce discrepancies from the original image in terms of fine-grained details, and it struggles with effectively handling scenes involving motion, as well as MVSplat.
In contrast, the proposed method shows a clean and detailed output, especially in the area of depth information discontinuity.

\subsubsection{Stereo Video Synthesis}Moreover, we qualitatively compared the results of the stereo video synthesis with other methods. 
In addition to the quality issues of single-frame generation, the most important aspect in videos is the temporal consistency. We seek to generate videos that are free of flicker. The visualization of generated videos from different methods are illustrated in Figure~\ref{fig_video_compare_1}, for each video, we show the scanline (the yellow line in figure) slice through the spatial-temporal volume. Our method achieve the smoothest trajectories, while other methods exhibit significant flickering. The most distinct regions are zoomed in and highlighted under the images to illustrate the details. In the first case, our method attains the smooth trajectories of the background (maintaining the straight line) . And the second case, only our method preserves the texture of the trees behind the woman. And the last case, only our method preserves the texture of the black vertical lines.  More results of generated frames are also presented in supplementary material.

\subsection{Quantitative Results}
Table~\ref{tab:quan} lists the SSIM, PSNR, and LPIPS scores for the newly synthesized images. The proposed method outperforms  3D-photography and SynSin in the major metrics by a wide margin. 
The method of Wang et al., P-NVS and NVSVDE-Net perform slightly worse in terms of all metrics compared to the proposed method. Since the proposed method merges the spatial features and fills the occluded regions, it achieves the best LPIPS scores and shows competitive results in PSNR and SSIM when compared with AdaMPI and CoPoNeRF. MVSplat outperforms other methods when $t$=10 but is inferior to ours when $t$=5.
Overall, the results exhibit superior visual effects, which is consistent with the performance observed in the qualitative experiments.
We further demonstrate the effectiveness of our method on video synthesis, as shown in Table~\ref{tab:quan_2}. 
The proposed method achieves the best performance on both FVD and $E_{warp}$, which means that our generated videos attain the best quality and temporal consistency.

\begin{table}[t]
\centering
\tabcolsep=0.5mm
\caption{Quantitative comparison on RealEstate10K dataset. $\dagger$ means re-implementing in our setting for fair comparison. 
When $t$=5, the current frame serves as the left view and the frame at time$+5$ serves as the right view;  the same approach applies for $t$=10.
The best results are in \textbf{bold}.
}
\begin{tabular}{lcccccc}
\toprule
\multirow{2}{*}{} & \multicolumn{2}{c}{SSIM $\uparrow$} & \multicolumn{2}{c}{PSNR $\uparrow$} & \multicolumn{2}{c}{LPIPS $\downarrow$} \\
\cmidrule(lr){2-3} \cmidrule(lr){4-5} \cmidrule(lr){6-7}
 & $t$=5 & $t$=10 & $t$=5 & $t$=10 & $t$=5 & $t$=10\\
\midrule
 3D-photography~\cite{shih20203d}$\dagger$ & 0.641 & 0.446 & 11.08 & 9.533 & 0.116 & 0.386\\
SynSin~\cite{wiles2020synsin}$\dagger$ & 0.740 & 0.645 & 21.22 & 19.55 & 0.087 & 0.104\\
AdaMPI~\cite{han2022single}$\dagger$ & 0.869 & 0.736 & 27.70 & 22.65 & 0.041 & 0.087\\
SinMPI~\cite{pu2023sinmpinovelviewsynthesis}$\dagger$ & {0.898 } & {0.775} & {28.79} & {23.21} & 0.040 & 0.080\\
Wang et al.~\cite{wang2023learning} & 0.840 & 0.720 & 25.35 & 21.36 & 0.049 & 0.095\\
P-NVS~\cite{yu2023longterm}$\dagger$ & 0.724 & 0.667 & 21.98 & 19.46 &  0.241 & 0.279\\
NVSVDE-Net~\cite{bello2024novel} & \multicolumn{2}{c}{0.840} & \multicolumn{2}{c}{24.31} & \multicolumn{2}{c}{0.233}\\
CoPoNeRF~\cite{hong2024unifying}$\dagger$ & 0.658 &0.632 & 22.63 & 21.33 & 0.171 & 0.209\\
MVSplat~\cite{chen2024mvsplat}$\dagger$ & \multicolumn{2}{c}{0.863} & \multicolumn{2}{c}{25.89} & \multicolumn{2}{c}{0.132}\\
Proposed & \textbf{0.916} & \textbf{0.857} & \textbf{32.26} & \textbf{24.86} & \textbf{0.038} & \textbf{0.049}\\
\bottomrule
\end{tabular}
\vspace{-2mm}
\label{tab:quan}
\end{table}

\begin{table}[t]
\centering
\tabcolsep=3mm
\caption{Quantitative comparison with the other methods. The best results are in \textbf{bold}. $E^{*}_{warp}$ denotes $E_{warp}$ $(\times10^{-3})$.}
\begin{tabular}{lcc}
\toprule  
 &FVD $\downarrow$ & $E^{*}_{warp}$ $\downarrow$ \\
\midrule  
 3D-photography~\cite{shih20203d}& 155.0& 3.418\\
Webui-depthmap& 83.18& 3.486\\
P-NVS~\cite{yu2023longterm}& 420.2 & 23.02\\ 
CoPoNeRF~\cite{hong2024unifying}& 290.0&7.144 \\ 
NVS-Solver~\cite{you2024nvs}& 249.1 & 5.842\\
MVSplat~\cite{chen2024mvsplat}& 1543 & 7.913 \\
AVP~\cite{avp}& 99.92 & 3.446\\ 
Proposed& \textbf{67.09}& \textbf{3.374}\\
\bottomrule 
\end{tabular}
\label{tab:quan_2}
\end{table}

\begin{figure*}[t]
\centering
\includegraphics[width=0.95\textwidth]{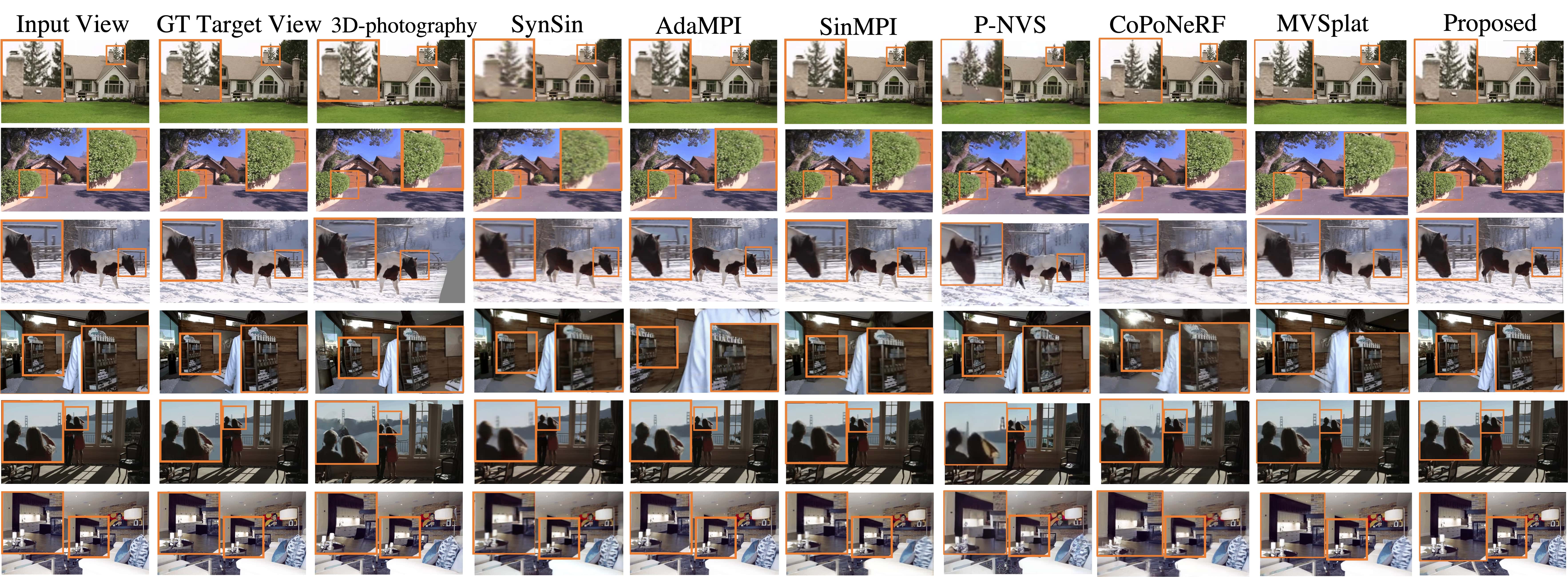}
\vspace{-2mm}
\caption{Qualitative results on the RealEstate10K dataset. The proposed method generates better-quality information and maintains the edge and textural details of the reference image better than the other methods. More results are available in supplementary material.}
\label{fig_compare_img}
\end{figure*}

\begin{figure*}[t]
    \centering
    \includegraphics[width=0.98\textwidth]{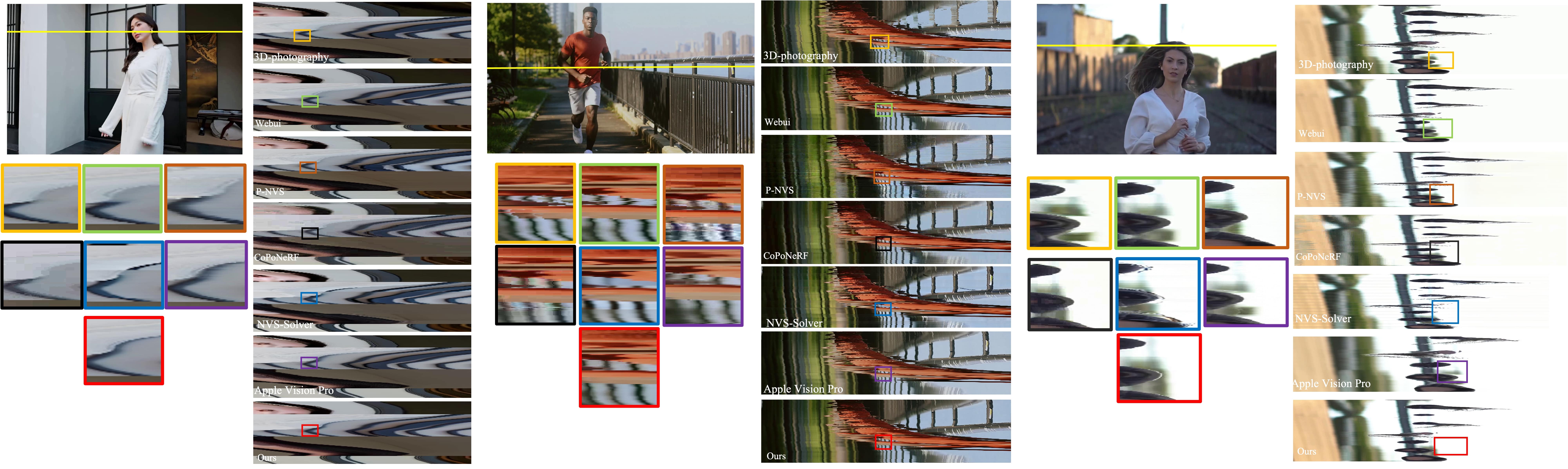}
    \vspace{-2mm}
    \caption{Visual comparisons in stereo video benchmark. We show the scanline (yellow line in original frame) slice through the spatial-temporal volume. The most distinct regions are zoomed in and highlighted under the images to illustrate the details.}
    \label{fig_video_compare_1}
\end{figure*}

\subsection{Ablation Study}

\begin{figure}[t]
    \centering
    \includegraphics[width=0.95\linewidth]{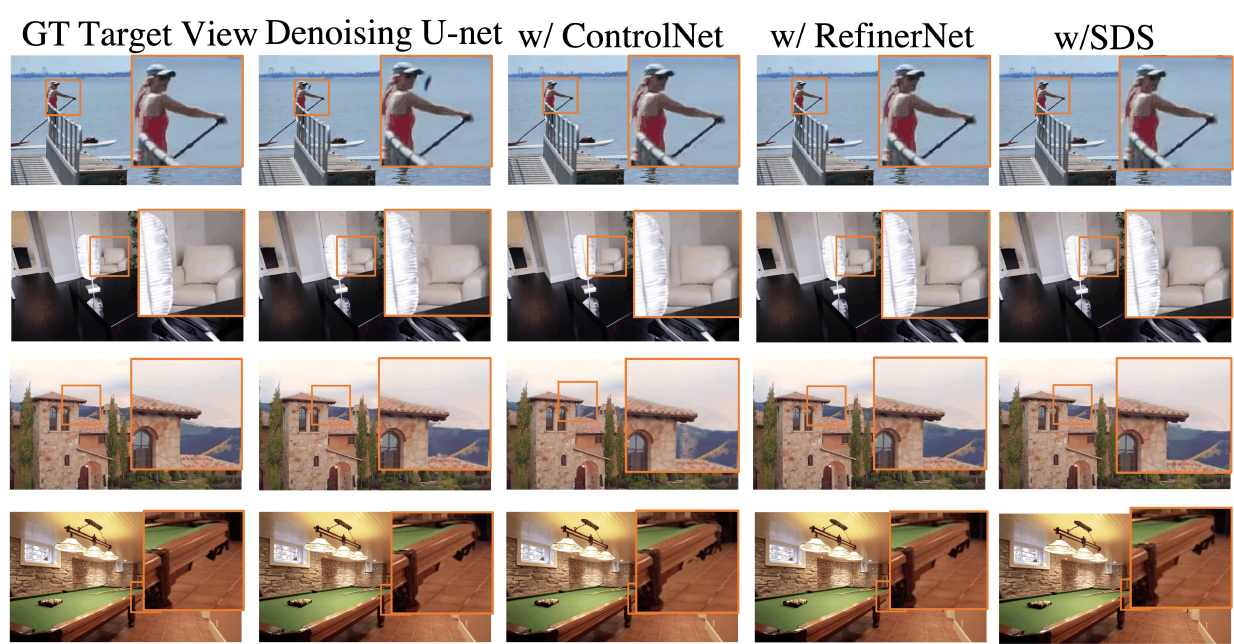}
    \caption{Ablation study of different designs. Only the proposed method ensures consistent preservation of details with the ground-truth target view. More results are available in supplementary material.}
    \label{fig_aba}
\end{figure}
\begin{table}[t]
    \centering
    \setlength{\tabcolsep}{2.5pt}
    \caption{Quantitative comparison for the ablation study. The best results are in \textbf{bold}.}
    \resizebox{1.\columnwidth}{!}{
        \begin{tabular}{ll|ccccc|ccc|c}
        \toprule[0.15em]
        \multicolumn{2}{l|}{\multirow{2}{*}{\textbf{Method}}} & \multicolumn{1}{c}{\multirow{2}{*}{\textbf{CN}}} & \multicolumn{1}{c}{\multirow{2}{*}{\textbf{RN}}} & \multicolumn{1}{c}{\multirow{2}{*}{\textbf{SDS}}} & \multicolumn{1}{c}{\multirow{2}{*}{\textbf{TIL}}} & \multicolumn{1}{c|}{\multirow{2}{*}{\textbf{DVG}}} &
        \multicolumn{3}{c|}{\textbf{Stereo Image}} & \multicolumn{1}{c}{\textbf{Stereo Video}} \\
        \multicolumn{7}{c|}{} & SSIM$\uparrow$ & PSNR$\uparrow$ & LPIPS$\downarrow$
        & FVD$\downarrow$ \\
        \midrule[0.15em]
        \multicolumn{2}{l|}{\multirow{4}{*}{\textbf{Image-level}}} & & & &  &  & 0.880 & 23.73 & 0.06 &-\\ 
        & & \checkmark &  &  &  &  & 0.855 & 24.04 & 0.183&- \\
        & &  & \checkmark &  &  &  & 0.895 & 30.20 & 0.043&- \\
        & &  & \checkmark & \checkmark &  &  & \textbf{0.916} & \textbf{32.26} & \textbf{0.038}&- \\
        \midrule[0.1em]
        \multicolumn{2}{l|}{\multirow{3}{*}{\textbf{Video-level}}} &  & \checkmark & \checkmark & &  & -& -&- & 184.0\\
        & &  & \checkmark & \checkmark & \checkmark &- &- & -& -& 123.5 \\
        & &  & \checkmark & \checkmark &  & \checkmark & -& -&- & 85.21\\
        & &  & \checkmark & \checkmark & \checkmark & \checkmark &- &- & -& \textbf{67.09}\\
        \bottomrule[0.1em]
    \end{tabular}{}
    }
    \label{tab:quan_1}
\end{table}

\subsubsection{Effectiveness of each components}
We show the results in Table~\ref{tab:quan_1} on the RealEstate10K (stereo image) and collected stereo video datasets. CN means substituting our RefinerNet (RN) to ControlNet, SDS means the proposed stereo deviation strength. \textbf{For image-level comparison}, the results show that utilizing U-Net architecture only is not sufficient for stereo image synthesis and our RefinerNet is a better choice than ControlNet. Adding SDS further improves the geometric consistency. 
As visualized in Figure~\ref{fig_aba}, solely relying on U-Net features results in distortion in the edge area and struggles with generating content properly. It also fails to preserve textural details near a plane boundary. Adding ControlNet often results in content inconsistency, presumably due to the failure of the paired view feature integration. RefinerNet handles this problem, showing it is a better architecture for stereo image generation. SDS further improves the consistency with the input and contain a more detailed image structure. As for \textbf{video-level comparison}, the quantitative results are shown in Table~\ref{tab:quan_1}, adding TIL and DVG both attain the performance gain, and our final model achieves the best performance.


\begin{figure}[t]
    \centering    \includegraphics[width=0.97\linewidth]{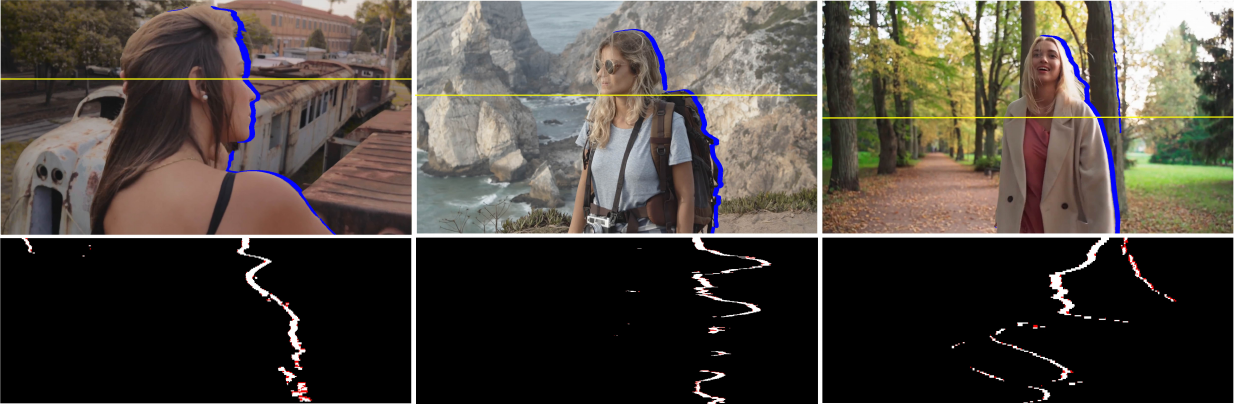}
    \caption{Visual comparisons for the improvement of stereo occlusion. We show the scanline of (yellow line in original frame) slice through the spatial-temporal volume. The red region represents the changes of occlusion mask.}
    \label{fig_dvg}
\end{figure}
\subsubsection{Data improvement by DVG}We show the generated stereo occlusion mask with inter-frame motion refinement in Figure~\ref{fig_dvg}. It can be observed that the occlusions appear more complete spatially and exhibit smoother in their temporal distribution. By employing DVG, we can acquire sufficient data to support the generation of spatio-temporally consistent videos. More analyses are available in the supplementary materials. 

\subsubsection{Exploration of different depth estimation methods}
We further compared different design choices of depth map including DepthAnything~\cite{depthanything}, Marigold~\cite{marigold}, MiDaS~\cite{Ranftl2022}, ZoeDepth~\cite{zoe} and DepthCrafter~\cite{hu2024-DepthCrafter}. 
The results are available in supplementary materials. The proposed method can produce new viewpoint images that are realistic, seamless, and rich in detail, regardless of the depth estimation method used. 
This capability demonstrates the versatility and robustness of the proposed method, ensuring superior performance in various applications and environments.

\subsection{Limitations}
SpatialDreamer effectively creates stereo video from a single input but faces limitations. It relies on depth maps for rendering, affecting stereoscopic quality due to depth accuracy and distribution.
Future research could focus on integrating implicit depth to address this~\cite{liang2024wonderland}. Additionally, the model's large parameter size limits real-time performance.
\section{Conclusion}

In this paper, we introduce a new self-supervised stereo video synthesis approach using a video diffusion model, termed SpatialDreamer. This method addresses the issues of insufficient data and spatio-temporal inconsistency between frames. To solve the problem of insufficient data, we develop a Depth-based Video Generation (DVG) module that uses a forward-backward rendering mechanism to generate a rendered video with geometric and temporal priors. Additionally, we propose RefinerNet along with a self-supervised synthetic framework to enable efficient and dedicated training using data generated by DVG. Furthermore, we design a consistency control module to ensure geometric and temporal consistency. Our SpatialDreamer outperforms all other open-source stereo image and video synthesis methods and has the potential for future expansion into virtual reality applications.

{
    \small
    \bibliographystyle{ieeenat_fullname}
    \bibliography{main}

\begin{thebibliography}{78}
\providecommand{\natexlab}[1]{#1}
\providecommand{\url}[1]{\texttt{#1}}
\expandafter\ifx\csname urlstyle\endcsname\relax
  \providecommand{\doi}[1]{doi: #1}\else
  \providecommand{\doi}{doi: \begingroup \urlstyle{rm}\Url}\fi

\bibitem[Akimov et~al.(2012)Akimov, Shestov, Voronov, and Vatolin]{occ_refine}
Dmitry Akimov, Alexey Shestov, Alexander Voronov, and Dmitriy Vatolin.
\newblock Occlusion refinement for stereo video using optical flow.
\newblock In \emph{2012 International Conference on 3D Imaging (IC3D)}, pages 1--8, 2012.

\bibitem[{Andrea Schubert}(2024)]{wwdc}
{Andrea Schubert}.
\newblock visionos 2 brings new spatial computing experiences to apple vision pro.
\newblock \url{https://www.apple.com/newsroom/2024/06/visionos-2-brings-new-spatial-computing-experiences-to-apple-vision-pro}, 2024.

\bibitem[Bar-Tal et~al.(2022)Bar-Tal, Ofri-Amar, Fridman, Kasten, and Dekel]{bar2022text2live}
Omer Bar-Tal, Dolev Ofri-Amar, Rafail Fridman, Yoni Kasten, and Tali Dekel.
\newblock Text2live: Text-driven layered image and video editing.
\newblock In \emph{European conference on computer vision}, pages 707--723. Springer, 2022.

\bibitem[Bello and Kim(2024)]{bello2024novel}
Juan Luis~Gonzalez Bello and Munchurl Kim.
\newblock Novel view synthesis with view-dependent effects from a single image.
\newblock In \emph{Proceedings of the IEEE/CVF Conference on Computer Vision and Pattern Recognition}, 2024.

\bibitem[Bhat et~al.(2023)Bhat, Birkl, Wofk, Wonka, and Müller]{zoe}
Shariq~Farooq Bhat, Reiner Birkl, Diana Wofk, Peter Wonka, and Matthias Müller.
\newblock Zoedepth: Zero-shot transfer by combining relative and metric depth, 2023.

\bibitem[Binh~Do and Chi~Nguyen(2019)]{areview}
Phuong~Ngoc Binh~Do and Quoc Chi~Nguyen.
\newblock A review of stereo-photogrammetry method for 3-d reconstruction in computer vision.
\newblock In \emph{2019 19th International Symposium on Communications and Information Technologies (ISCIT)}, pages 138--143, 2019.

\bibitem[Blattmann et~al.(2023{\natexlab{a}})Blattmann, Dockhorn, Kulal, Mendelevitch, Kilian, Lorenz, Levi, English, Voleti, Letts, et~al.]{blattmann2023stable}
Andreas Blattmann, Tim Dockhorn, Sumith Kulal, Daniel Mendelevitch, Maciej Kilian, Dominik Lorenz, Yam Levi, Zion English, Vikram Voleti, Adam Letts, et~al.
\newblock Stable video diffusion: Scaling latent video diffusion models to large datasets.
\newblock \emph{arXiv preprint arXiv:2311.15127}, 2023{\natexlab{a}}.

\bibitem[Blattmann et~al.(2023{\natexlab{b}})Blattmann, Rombach, Ling, Dockhorn, Kim, Fidler, and Kreis]{blattmann2023align}
Andreas Blattmann, Robin Rombach, Huan Ling, Tim Dockhorn, Seung~Wook Kim, Sanja Fidler, and Karsten Kreis.
\newblock Align your latents: High-resolution video synthesis with latent diffusion models.
\newblock In \emph{Proceedings of the IEEE/CVF Conference on Computer Vision and Pattern Recognition}, pages 22563--22575, 2023{\natexlab{b}}.

\bibitem[Bobick and Intille(1999)]{LargeOcclusionStereo}
Aaron Bobick and Stephen Intille.
\newblock Large occlusion stereo.
\newblock \emph{International Journal of Computer Vision}, 33:\penalty0 181--200, 1999.

\bibitem[Brooks et~al.(2024)Brooks, Peebles, Holmes, DePue, Guo, Jing, Schnurr, Taylor, Luhman, Luhman, Ng, Wang, and Ramesh]{videoworldsimulators2024}
Tim Brooks, Bill Peebles, Connor Holmes, Will DePue, Yufei Guo, Li Jing, David Schnurr, Joe Taylor, Troy Luhman, Eric Luhman, Clarence Ng, Ricky Wang, and Aditya Ramesh.
\newblock Video generation models as world simulators.
\newblock 2024.

\bibitem[Cao et~al.(2023)Cao, Wang, Qi, Shan, Qie, and Zheng]{cao2023masactrltuning}
Mingdeng Cao, Xintao Wang, Zhongang Qi, Ying Shan, Xiaohu Qie, and Yinqiang Zheng.
\newblock Masactrl: Tuning-free mutual self-attention control for consistent image synthesis and editing, 2023.

\bibitem[Charatan et~al.(2023)Charatan, Li, Tagliasacchi, and Sitzmann]{charatan23pixelsplat}
David Charatan, Sizhe Li, Andrea Tagliasacchi, and Vincent Sitzmann.
\newblock pixelsplat: 3d gaussian splats from image pairs for scalable generalizable 3d reconstruction.
\newblock In \emph{arXiv}, 2023.

\bibitem[Chen et~al.(2023)Chen, Wu, Xie, Wu, Li, Xia, Xiao, and Lin]{chen2023control}
Weifeng Chen, Jie Wu, Pan Xie, Hefeng Wu, Jiashi Li, Xin Xia, Xuefeng Xiao, and Liang Lin.
\newblock Control-a-video: Controllable text-to-video generation with diffusion models.
\newblock \emph{arXiv preprint arXiv:2305.13840}, 2023.

\bibitem[Chen et~al.(2024)Chen, Xu, Zheng, Zhuang, Pollefeys, Geiger, Cham, and Cai]{chen2024mvsplat}
Yuedong Chen, Haofei Xu, Chuanxia Zheng, Bohan Zhuang, Marc Pollefeys, Andreas Geiger, Tat-Jen Cham, and Jianfei Cai.
\newblock Mvsplat: Efficient 3d gaussian splatting from sparse multi-view images.
\newblock \emph{arXiv preprint arXiv:2403.14627}, 2024.

\bibitem[Dai et~al.(2023)Dai, Zhang, Yao, Qiu, Zhu, Qin, and Wang]{dai2023animateanything}
Zuozhuo Dai, Zhenghao Zhang, Yao Yao, Bingxue Qiu, Siyu Zhu, Long Qin, and Weizhi Wang.
\newblock Animateanything: Fine-grained open domain image animation with motion guidance.
\newblock \emph{arXiv e-prints}, pages arXiv--2311, 2023.

\bibitem[Egnal and Wildes(2002)]{1023808}
G. Egnal and R.P. Wildes.
\newblock Detecting binocular half-occlusions: empirical comparisons of five approaches.
\newblock \emph{IEEE Transactions on Pattern Analysis and Machine Intelligence}, 24\penalty0 (8):\penalty0 1127--1133, 2002.

\bibitem[Esser et~al.(2023)Esser, Chiu, Atighehchian, Granskog, and Germanidis]{esser2023structure}
Patrick Esser, Johnathan Chiu, Parmida Atighehchian, Jonathan Granskog, and Anastasis Germanidis.
\newblock Structure and content-guided video synthesis with diffusion models.
\newblock In \emph{Proceedings of the IEEE/CVF International Conference on Computer Vision}, pages 7346--7356, 2023.

\bibitem[Gao et~al.(2021)Gao, Saraf, Kopf, and Huang]{gao2021dynamicviewsynthesisdynamic}
Chen Gao, Ayush Saraf, Johannes Kopf, and Jia-Bin Huang.
\newblock Dynamic view synthesis from dynamic monocular video, 2021.

\bibitem[Guo et~al.(2024)Guo, Yang, Rao, Liang, Wang, Qiao, Agrawala, Lin, and Dai]{guo2023animatediff}
Yuwei Guo, Ceyuan Yang, Anyi Rao, Zhengyang Liang, Yaohui Wang, Yu Qiao, Maneesh Agrawala, Dahua Lin, and Bo Dai.
\newblock Animatediff: Animate your personalized text-to-image diffusion models without specific tuning.
\newblock \emph{International Conference on Learning Representations}, 2024.

\bibitem[Gupta et~al.(2023)Gupta, Yu, Sohn, Gu, Hahn, Fei-Fei, Essa, Jiang, and Lezama]{gupta2023photorealistic}
Agrim Gupta, Lijun Yu, Kihyuk Sohn, Xiuye Gu, Meera Hahn, Li Fei-Fei, Irfan Essa, Lu Jiang, and Jos{\'e} Lezama.
\newblock Photorealistic video generation with diffusion models.
\newblock \emph{arXiv preprint arXiv:2312.06662}, 2023.

\bibitem[Han et~al.(2022)Han, Wang, and Yang]{han2022single}
Yuxuan Han, Ruicheng Wang, and Jiaolong Yang.
\newblock Single-view view synthesis in the wild with learned adaptive multiplane images.
\newblock In \emph{ACM SIGGRAPH 2022 Conference Proceedings}, pages 1--8, 2022.

\bibitem[Hertz et~al.(2022)Hertz, Mokady, Tenenbaum, Aberman, Pritch, and Cohen-Or]{hertz2022imageeditingcross}
Amir Hertz, Ron Mokady, Jay Tenenbaum, Kfir Aberman, Yael Pritch, and Daniel Cohen-Or.
\newblock Prompt-to-prompt image editing with cross attention control, 2022.

\bibitem[Ho et~al.(2022)Ho, Salimans, Gritsenko, Chan, Norouzi, and Fleet]{ho2022video}
Jonathan Ho, Tim Salimans, Alexey Gritsenko, William Chan, Mohammad Norouzi, and David~J Fleet.
\newblock Video diffusion models.
\newblock In \emph{Advances in Neural Information Processing Systems}. Curran Associates, Inc., 2022.

\bibitem[Hong et~al.(2024)Hong, Jung, Shin, Yang, Kim, and Luo]{hong2024unifying}
Sunghwan Hong, Jaewoo Jung, Heeseong Shin, Jiaolong Yang, Seungryong Kim, and Chong Luo.
\newblock Unifying correspondence, pose and nerf for pose-free novel view synthesis from stereo pairs, 2024.

\bibitem[Hu et~al.(2023)Hu, Gao, Zhang, Sun, Zhang, and Bo]{hu2023animate}
Li Hu, Xin Gao, Peng Zhang, Ke Sun, Bang Zhang, and Liefeng Bo.
\newblock Animate anyone: Consistent and controllable image-to-video synthesis for character animation.
\newblock \emph{arXiv preprint arXiv:2311.17117}, 2023.

\bibitem[Hu et~al.(2024)Hu, Gao, Li, Zhao, Cun, Zhang, Quan, and Shan]{hu2024-DepthCrafter}
Wenbo Hu, Xiangjun Gao, Xiaoyu Li, Sijie Zhao, Xiaodong Cun, Yong Zhang, Long Quan, and Ying Shan.
\newblock Depthcrafter: Generating consistent long depth sequences for open-world videos.
\newblock \emph{arXiv preprint arXiv:2409.02095}, 2024.

\bibitem[Jampani et~al.(2021)Jampani, Chang, Sargent, Kar, Tucker, Krainin, Kaeser, Freeman, Salesin, Curless, et~al.]{jampani2021slide}
Varun Jampani, Huiwen Chang, Kyle Sargent, Abhishek Kar, Richard Tucker, Michael Krainin, Dominik Kaeser, William~T Freeman, David Salesin, Brian Curless, et~al.
\newblock Slide: Single image 3d photography with soft layering and depth-aware inpainting.
\newblock In \emph{Proceedings of the IEEE/CVF International Conference on Computer Vision}, pages 12518--12527, 2021.

\bibitem[Jang and Agapito(2024)]{jang2024nvistwildnewview}
Wonbong Jang and Lourdes Agapito.
\newblock Nvist: In the wild new view synthesis from a single image with transformers, 2024.

\bibitem[Jeong et~al.(2024)Jeong, Cai, Garrepalli, Lin, Hayat, and Porikli]{jeong2024ocaiimprovingopticalflow}
Jisoo Jeong, Hong Cai, Risheek Garrepalli, Jamie~Menjay Lin, Munawar Hayat, and Fatih Porikli.
\newblock Ocai: Improving optical flow estimation by occlusion and consistency aware interpolation, 2024.

\bibitem[Jin et~al.(2005)Jin, Soatto, and Yezzi]{jin2005multi}
Hailin Jin, Stefano Soatto, and Anthony~J Yezzi.
\newblock Multi-view stereo reconstruction of dense shape and complex appearance.
\newblock \emph{International Journal of Computer Vision}, 63:\penalty0 175--189, 2005.

\bibitem[Ke et~al.(2024)Ke, Obukhov, Huang, Metzger, Daudt, and Schindler]{marigold}
Bingxin Ke, Anton Obukhov, Shengyu Huang, Nando Metzger, Rodrigo~Caye Daudt, and Konrad Schindler.
\newblock Repurposing diffusion-based image generators for monocular depth estimation, 2024.

\bibitem[Kerbl et~al.(2023)Kerbl, Kopanas, Leimkühler, and Drettakis]{kerbl20233dgaussiansplattingrealtime}
Bernhard Kerbl, Georgios Kopanas, Thomas Leimkühler, and George Drettakis.
\newblock 3d gaussian splatting for real-time radiance field rendering, 2023.

\bibitem[Khachatryan et~al.(2023)Khachatryan, Movsisyan, Tadevosyan, Henschel, Wang, Navasardyan, and Shi]{khachatryan2023text2video}
Levon Khachatryan, Andranik Movsisyan, Vahram Tadevosyan, Roberto Henschel, Zhangyang Wang, Shant Navasardyan, and Humphrey Shi.
\newblock Text2video-zero: Text-to-image diffusion models are zero-shot video generators.
\newblock \emph{arXiv preprint arXiv:2303.13439}, 2023.

\bibitem[Kondratyuk et~al.(2023)Kondratyuk, Yu, Gu, Lezama, Huang, Hornung, Adam, Akbari, Alon, Birodkar, et~al.]{kondratyuk2023videopoet}
Dan Kondratyuk, Lijun Yu, Xiuye Gu, Jos{\'e} Lezama, Jonathan Huang, Rachel Hornung, Hartwig Adam, Hassan Akbari, Yair Alon, Vighnesh Birodkar, et~al.
\newblock Videopoet: A large language model for zero-shot video generation.
\newblock \emph{arXiv preprint arXiv:2312.14125}, 2023.

\bibitem[Lai et~al.(2018)Lai, Huang, Wang, Shechtman, Yumer, and Yang]{lai2018learning}
Wei-Sheng Lai, Jia-Bin Huang, Oliver Wang, Eli Shechtman, Ersin Yumer, and Ming-Hsuan Yang.
\newblock Learning blind video temporal consistency.
\newblock In \emph{Proceedings of the European conference on computer vision (ECCV)}, pages 170--185, 2018.

\bibitem[Liang et~al.(2024)Liang, Cao, Goel, Qian, Korolev, Terzopoulos, Plataniotis, Tulyakov, and Ren]{liang2024wonderland}
Hanwen Liang, Junli Cao, Vidit Goel, Guocheng Qian, Sergei Korolev, Demetri Terzopoulos, Konstantinos Plataniotis, Sergey Tulyakov, and Jian Ren.
\newblock Wonderland: Navigating 3d scenes from a single image.
\newblock \emph{arXiv preprint arXiv:2412.12091}, 2024.

\bibitem[Liu et~al.(2021)Liu, Tucker, Jampani, Makadia, Snavely, and Kanazawa]{liu2021infinite}
Andrew Liu, Richard Tucker, Varun Jampani, Ameesh Makadia, Noah Snavely, and Angjoo Kanazawa.
\newblock Infinite nature: Perpetual view generation of natural scenes from a single image.
\newblock In \emph{Proceedings of the IEEE/CVF International Conference on Computer Vision}, 2021.

\bibitem[Liu et~al.(2024)Liu, Zhang, Li, Yan, Gao, Chen, Yuan, Huang, Sun, Gao, He, and Sun]{liu2024sorareviewbackgroundtechnology}
Yixin Liu, Kai Zhang, Yuan Li, Zhiling Yan, Chujie Gao, Ruoxi Chen, Zhengqing Yuan, Yue Huang, Hanchi Sun, Jianfeng Gao, Lifang He, and Lichao Sun.
\newblock Sora: A review on background, technology, limitations, and opportunities of large vision models, 2024.

\bibitem[Loshchilov and Hutter(2017)]{loshchilov2017decoupled}
Ilya Loshchilov and Frank Hutter.
\newblock Decoupled weight decay regularization.
\newblock \emph{arXiv preprint arXiv:1711.05101}, 2017.

\bibitem[Matzen et~al.(2017)Matzen, Cohen, Evans, Kopf, and Szeliski]{Low-cost360}
Kevin Matzen, Michael~F. Cohen, Bryce Evans, Johannes Kopf, and Richard Szeliski.
\newblock Low-cost 360 stereo photography and video capture.
\newblock \emph{ACM Trans. Graph.}, 36\penalty0 (4), 2017.

\bibitem[Mildenhall et~al.(2020)Mildenhall, Srinivasan, Tancik, Barron, Ramamoorthi, and Ng]{mildenhall2020nerfrepresentingscenesneural}
Ben Mildenhall, Pratul~P. Srinivasan, Matthew Tancik, Jonathan~T. Barron, Ravi Ramamoorthi, and Ren Ng.
\newblock Nerf: Representing scenes as neural radiance fields for view synthesis, 2020.

\bibitem[M{\"u}ller et~al.(2024)M{\"u}ller, Schwarz, R{\"o}ssle, Porzi, Bul{\`o}, Nie{\ss}ner, and Kontschieder]{muller2024multidiff}
Norman M{\"u}ller, Katja Schwarz, Barbara R{\"o}ssle, Lorenzo Porzi, Samuel~Rota Bul{\`o}, Matthias Nie{\ss}ner, and Peter Kontschieder.
\newblock Multidiff: Consistent novel view synthesis from a single image.
\newblock In \emph{Proceedings of the IEEE/CVF Conference on Computer Vision and Pattern Recognition}, pages 10258--10268, 2024.

\bibitem[Niklaus and Liu(2020)]{niklaus2020softmaxsplattingvideoframe}
Simon Niklaus and Feng Liu.
\newblock Softmax splatting for video frame interpolation, 2020.

\bibitem[{Onee}(2024)]{avp}
{Onee}.
\newblock What is spatial video on iphone 15 pro and vision pro.
\newblock \url{https://xreality.zone/en/posts/what-is-spatial-video-on-iphone-15-pro-and-apple-vision-pro}, 2024.

\bibitem[Ouyang et~al.(2023)Ouyang, Wang, Xiao, Bai, Zhang, Zheng, Zhou, Chen, and Shen]{ouyang2023codef}
Hao Ouyang, Qiuyu Wang, Yuxi Xiao, Qingyan Bai, Juntao Zhang, Kecheng Zheng, Xiaowei Zhou, Qifeng Chen, and Yujun Shen.
\newblock Codef: Content deformation fields for temporally consistent video processing.
\newblock \emph{arXiv preprint arXiv:2308.07926}, 2023.

\bibitem[{\"O}zye{\c{s}}il et~al.(2017){\"O}zye{\c{s}}il, Voroninski, Basri, and Singer]{ozyecsil2017survey}
Onur {\"O}zye{\c{s}}il, Vladislav Voroninski, Ronen Basri, and Amit Singer.
\newblock A survey of structure from motion*.
\newblock \emph{Acta Numerica}, 26:\penalty0 305--364, 2017.

\bibitem[Pu et~al.(2023)Pu, Wang, and Lian]{pu2023sinmpinovelviewsynthesis}
Guo Pu, Peng-Shuai Wang, and Zhouhui Lian.
\newblock Sinmpi: Novel view synthesis from a single image with expanded multiplane images, 2023.

\bibitem[Qi et~al.(2023)Qi, Cun, Zhang, Lei, Wang, Shan, and Chen]{qi2023fatezero}
Chenyang Qi, Xiaodong Cun, Yong Zhang, Chenyang Lei, Xintao Wang, Ying Shan, and Qifeng Chen.
\newblock Fatezero: Fusing attentions for zero-shot text-based video editing.
\newblock \emph{arXiv preprint arXiv:2303.09535}, 2023.

\bibitem[Ranftl et~al.(2022)Ranftl, Lasinger, Hafner, Schindler, and Koltun]{Ranftl2022}
Ren\'{e} Ranftl, Katrin Lasinger, David Hafner, Konrad Schindler, and Vladlen Koltun.
\newblock Towards robust monocular depth estimation: Mixing datasets for zero-shot cross-dataset transfer.
\newblock \emph{IEEE Transactions on Pattern Analysis and Machine Intelligence}, 44\penalty0 (3), 2022.

\bibitem[Rombach et~al.(2022)Rombach, Blattmann, Lorenz, Esser, and Ommer]{rombach2022high}
Robin Rombach, Andreas Blattmann, Dominik Lorenz, Patrick Esser, and Bj{\"o}rn Ommer.
\newblock High-resolution image synthesis with latent diffusion models.
\newblock In \emph{Proceedings of the IEEE/CVF conference on computer vision and pattern recognition}, pages 10684--10695, 2022.

\bibitem[Sager et~al.(2021)Sager, Janiesch, and Zschech]{imagelabelling}
Christoph Sager, Christian Janiesch, and Patrick Zschech.
\newblock A survey of image labelling for computer vision applications.
\newblock \emph{Journal of Business Analytics}, 4\penalty0 (2):\penalty0 91–110, 2021.

\bibitem[Scharstein(1999)]{ViewSynthesis}
D. Scharstein.
\newblock \emph{View Synthesis Using Stereo Vision}.
\newblock Springer Berlin Heidelberg, 1999.

\bibitem[Schonberger and Frahm(2016)]{schonberger2016structure}
Johannes~L Schonberger and Jan-Michael Frahm.
\newblock Structure-from-motion revisited.
\newblock In \emph{Proceedings of the IEEE conference on computer vision and pattern recognition}, pages 4104--4113, 2016.

\bibitem[Seitz et~al.(2006)Seitz, Curless, Diebel, Scharstein, and Szeliski]{1640800}
S.M. Seitz, B. Curless, J. Diebel, D. Scharstein, and R. Szeliski.
\newblock A comparison and evaluation of multi-view stereo reconstruction algorithms.
\newblock In \emph{2006 IEEE Computer Society Conference on Computer Vision and Pattern Recognition (CVPR'06)}, pages 519--528, 2006.

\bibitem[Shi et~al.(2021)Shi, Li, and Yu]{shi2021selfsupervised}
Yujiao Shi, Hongdong Li, and Xin Yu.
\newblock Self-supervised visibility learning for novel view synthesis.*.
\newblock In \emph{Proceedings of the IEEE Conference on Computer Vision and Pattern Recognition}, 2021.

\bibitem[Shih et~al.(2020)Shih, Su, Kopf, and Huang]{shih20203d}
Meng-Li Shih, Shih-Yang Su, Johannes Kopf, and Jia-Bin Huang.
\newblock 3d photography using context-aware layered depth inpainting.
\newblock In \emph{Proceedings of the IEEE/CVF Conference on Computer Vision and Pattern Recognition}, pages 8028--8038, 2020.

\bibitem[Singer et~al.(2022)Singer, Polyak, Hayes, Yin, An, Zhang, Hu, Yang, Ashual, Gafni, et~al.]{singer2022make}
Uriel Singer, Adam Polyak, Thomas Hayes, Xi Yin, Jie An, Songyang Zhang, Qiyuan Hu, Harry Yang, Oron Ashual, Oran Gafni, et~al.
\newblock Make-a-video: Text-to-video generation without text-video data.
\newblock \emph{arXiv preprint arXiv:2209.14792}, 2022.

\bibitem[Stacchio(2023)]{stacchio2023stableinpainting}
Lorenzo Stacchio.
\newblock Train stable diffusion for inpainting, 2023.

\bibitem[Sun et~al.(2010)Sun, Xu, Au, Chui, and Kwok]{DIBR}
Wenxiu Sun, Lingfeng Xu, Oscar~C Au, Sung~Him Chui, and Chun~Wing Kwok.
\newblock An overview of free view-point depth-image-based rendering (dibr).
\newblock In \emph{APSIPA Annual Summit and Conference}, pages 1023--1030, 2010.

\bibitem[Teed and Deng(2020)]{raft}
Zachary Teed and Jia Deng.
\newblock Raft: Recurrent all-pairs field transforms for optical flow, 2020.

\bibitem[Tucker and Snavely(2020)]{tucker2020single}
Richard Tucker and Noah Snavely.
\newblock Single-view view synthesis with multiplane images.
\newblock In \emph{Proceedings of the IEEE/CVF Conference on Computer Vision and Pattern Recognition}, pages 551--560, 2020.

\bibitem[Unterthiner et~al.(2019)Unterthiner, van Steenkiste, Kurach, Marinier, Michalski, and Gelly]{DBLP:conf/iclr/UnterthinerSKMM19}
Thomas Unterthiner, Sjoerd van Steenkiste, Karol Kurach, Rapha{\"{e}}l Marinier, Marcin Michalski, and Sylvain Gelly.
\newblock {FVD:} {A} new metric for video generation.
\newblock In \emph{Deep Generative Models for Highly Structured Data, {ICLR} 2019 Workshop, New Orleans, Louisiana, United States, May 6, 2019}. OpenReview.net, 2019.

\bibitem[Van~Hoorick et~al.(2024)Van~Hoorick, Wu, Ozguroglu, Sargent, Liu, Tokmakov, Dave, Zheng, and Vondrick]{vanhoorick2024gcd}
Basile Van~Hoorick, Rundi Wu, Ege Ozguroglu, Kyle Sargent, Ruoshi Liu, Pavel Tokmakov, Achal Dave, Changxi Zheng, and Carl Vondrick.
\newblock Generative camera dolly: Extreme monocular dynamic novel view synthesis.
\newblock \emph{European Conference on Computer Vision (ECCV)}, 2024.

\bibitem[Vaswani et~al.(2017)Vaswani, Shazeer, Parmar, Uszkoreit, Jones, Gomez, Kaiser, and Polosukhin]{vaswani2017attention}
Ashish Vaswani, Noam Shazeer, Niki Parmar, Jakob Uszkoreit, Llion Jones, Aidan~N Gomez, {\L}ukasz Kaiser, and Illia Polosukhin.
\newblock Attention is all you need.
\newblock \emph{Advances in neural information processing systems}, 30, 2017.

\bibitem[Wang et~al.(2023{\natexlab{a}})Wang, Jiang, Xie, Liu, Chen, Cao, Wang, and Shen]{wang2023zero}
Wen Wang, Yan Jiang, Kangyang Xie, Zide Liu, Hao Chen, Yue Cao, Xinlong Wang, and Chunhua Shen.
\newblock Zero-shot video editing using off-the-shelf image diffusion models.
\newblock \emph{arXiv preprint arXiv:2303.17599}, 2023{\natexlab{a}}.

\bibitem[Wang et~al.(2023{\natexlab{b}})Wang, Wu, Yin, Ni, Wang, Li, Yang, Yang, Wang, Liu, et~al.]{wang2023learning}
Xiaodong Wang, Chenfei Wu, Shengming Yin, Minheng Ni, Jianfeng Wang, Linjie Li, Zhengyuan Yang, Fan Yang, Lijuan Wang, Zicheng Liu, et~al.
\newblock Learning 3d photography videos via self-supervised diffusion on single images.
\newblock \emph{arXiv preprint arXiv:2302.10781}, 2023{\natexlab{b}}.

\bibitem[Wang et~al.(2024)Wang, Yuan, Zhang, Chen, Wang, Zhang, Shen, Zhao, and Zhou]{wang2024videocomposer}
Xiang Wang, Hangjie Yuan, Shiwei Zhang, Dayou Chen, Jiuniu Wang, Yingya Zhang, Yujun Shen, Deli Zhao, and Jingren Zhou.
\newblock Videocomposer: Compositional video synthesis with motion controllability.
\newblock \emph{Advances in Neural Information Processing Systems}, 36, 2024.

\bibitem[Weng et~al.(2019)Weng, Curless, and Kemelmacher-Shlizerman]{weng2019photo}
Chung-Yi Weng, Brian Curless, and Ira Kemelmacher-Shlizerman.
\newblock Photo wake-up: 3d character animation from a single photo.
\newblock In \emph{Proceedings of the IEEE/CVF conference on computer vision and pattern recognition}, pages 5908--5917, 2019.

\bibitem[Wiles et~al.(2020)Wiles, Gkioxari, Szeliski, and Johnson]{wiles2020synsin}
Olivia Wiles, Georgia Gkioxari, Richard Szeliski, and Justin Johnson.
\newblock Synsin: End-to-end view synthesis from a single image.
\newblock In \emph{Proceedings of the IEEE/CVF Conference on Computer Vision and Pattern Recognition}, 2020.

\bibitem[Woods et~al.(2002)Woods, Docherty, and Koch]{imageDistortions}
Andrew Woods, Tom Docherty, and Rolf Koch.
\newblock Image distortions in stereoscopic video systems.
\newblock \emph{Proc SPIE}, 1915, 2002.

\bibitem[Wu et~al.(2023)Wu, Ge, Wang, Lei, Gu, Shi, Hsu, Shan, Qie, and Shou]{wu2023tune}
Jay~Zhangjie Wu, Yixiao Ge, Xintao Wang, Stan~Weixian Lei, Yuchao Gu, Yufei Shi, Wynne Hsu, Ying Shan, Xiaohu Qie, and Mike~Zheng Shou.
\newblock Tune-a-video: One-shot tuning of image diffusion models for text-to-video generation.
\newblock In \emph{Proceedings of the IEEE/CVF International Conference on Computer Vision}, pages 7623--7633, 2023.

\bibitem[Xu et~al.(2022)Xu, Jiang, Wang, Fan, Shi, and Wang]{xu2022sinnerftrainingneuralradiance}
Dejia Xu, Yifan Jiang, Peihao Wang, Zhiwen Fan, Humphrey Shi, and Zhangyang Wang.
\newblock Sinnerf: Training neural radiance fields on complex scenes from a single image, 2022.

\bibitem[Yang et~al.(2024)Yang, Kang, Huang, Xu, Feng, and Zhao]{depthanything}
Lihe Yang, Bingyi Kang, Zilong Huang, Xiaogang Xu, Jiashi Feng, and Hengshuang Zhao.
\newblock Depth anything: Unleashing the power of large-scale unlabeled data, 2024.

\bibitem[You et~al.(2024)You, Zhu, Liu, and Hou]{you2024nvs}
Meng You, Zhiyu Zhu, Hui Liu, and Junhui Hou.
\newblock Nvs-solver: Video diffusion model as zero-shot novel view synthesizer.
\newblock \emph{arXiv preprint arXiv:2405.15364}, 2024.

\bibitem[Yu et~al.(2023)Yu, Forghani, Derpanis, and Brubaker]{yu2023longterm}
Jason~J. Yu, Fereshteh Forghani, Konstantinos~G. Derpanis, and Marcus~A. Brubaker.
\newblock Long-term photometric consistent novel view synthesis with diffusion models, 2023.

\bibitem[Zhang et~al.(2007)Zhang, Hua, Qin, Wong, and Bao]{StereoscopicVideoSynthesis}
Guofeng Zhang, Wei Hua, Xueying Qin, Tien-Tsin Wong, and Hujun Bao.
\newblock Stereoscopic video synthesis from a monocular video.
\newblock \emph{IEEE Transactions on Visualization and Computer Graphics}, 13\penalty0 (4):\penalty0 686--696, 2007.

\bibitem[Zhao et~al.(2023)Zhao, Wang, Bao, Li, and Zhu]{zhao2023controlvideo}
Min Zhao, Rongzhen Wang, Fan Bao, Chongxuan Li, and Jun Zhu.
\newblock Controlvideo: Adding conditional control for one shot text-to-video editing.
\newblock \emph{arXiv preprint arXiv:2305.17098}, 2023.

\bibitem[Zhou et~al.(2018)Zhou, Tucker, Flynn, Fyffe, and Snavely]{zhou2018}
Tinghui Zhou, Richard Tucker, John Flynn, Graham Fyffe, and Noah Snavely.
\newblock Stereo magnification: Learning view synthesis using multiplane images.
\newblock \emph{CoRR}, abs/1805.09817, 2018.

\end{thebibliography}
}


\clearpage
\setcounter{page}{1}
\maketitlesupplementary

\section{Image Synthesis}

\subsection{Visual Comparison with Other Methods}
More results of comparison with other methods are shown in Figure~\ref{fig:img}.

\begin{figure*}[!ht]
    \vspace{-2cm}
    \centering
    \includegraphics[width=1.0\textwidth]{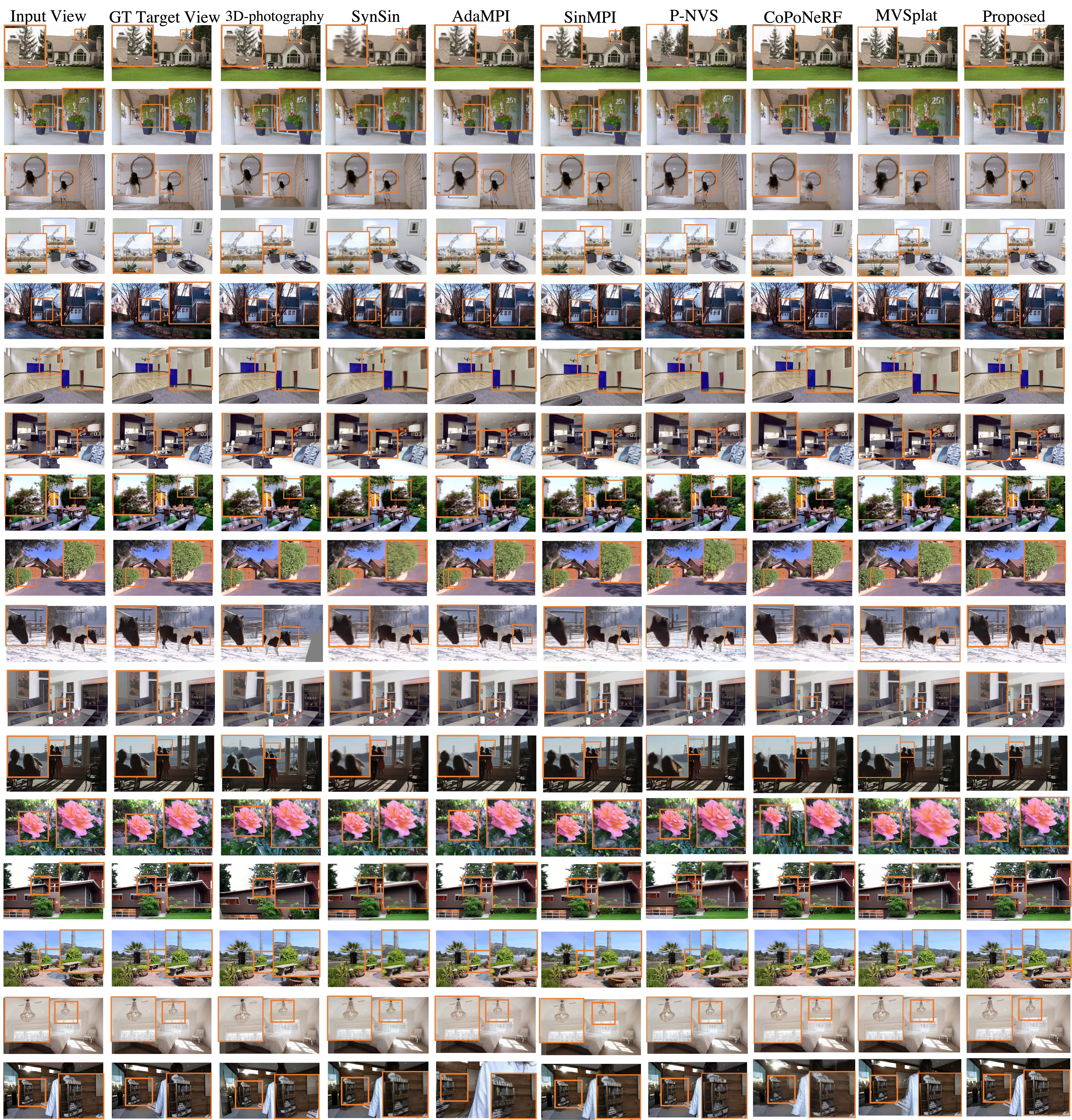}
    \caption{Visual result on the RealEstate10K dataset.}
    \label{fig:img}
\end{figure*}

\subsection{Visual Result of Ablation Study}
More results of ablation study are shown in Figure~\ref{fig:aba}. By zooming in on the images, one can more clearly discern the differences in detail, including artifacts, noise, and discontinuities.
\begin{figure*}[!ht]
    \vspace{-2cm}
    \centering
    \includegraphics[width=0.9\textwidth]{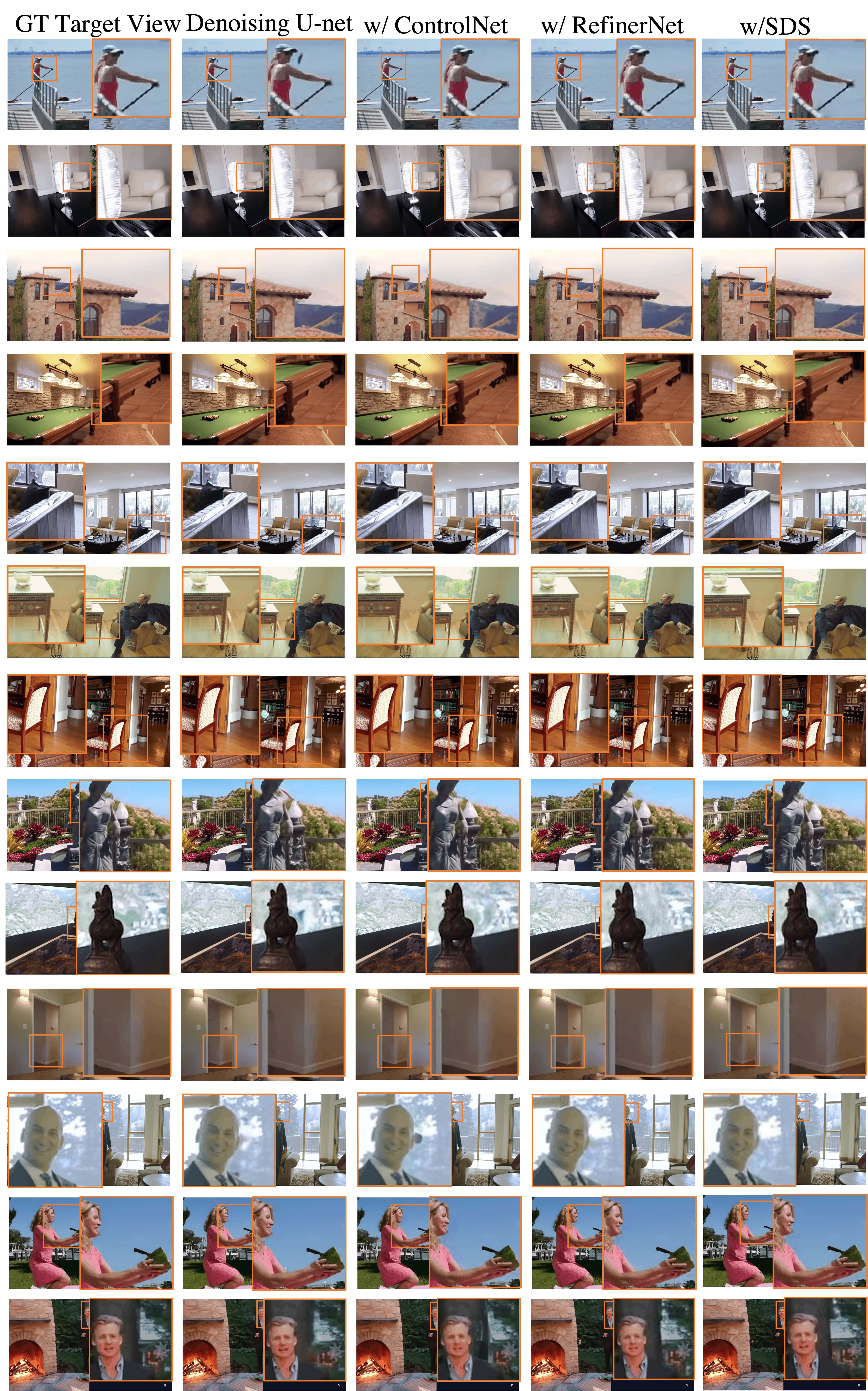}
    \caption{Visual result of ablation study.}
    \label{fig:aba}
\end{figure*}
       
\subsection{Visual Comparison of Different Depth Estimation Methods}
As shown in Figure~\ref{fig:depth}, DepthAnything is effective in capturing fine details and maintaining the consistency of depth across different scenes, while Marigold produces sharp and detailed depth maps. MidaS produces smooth and coherent depth maps, and ZoeDepth shows a balanced approach to detail preservation and depth accuracy, especially better reconstruction of flat surfaces. 
The proposed method consistently delivers excellent results in generating new viewpoint images, regardless of the depth estimation method used. 
\begin{figure*}[!ht]
    \vspace{-2cm}
    \centering
    \includegraphics[width=0.75\textwidth]{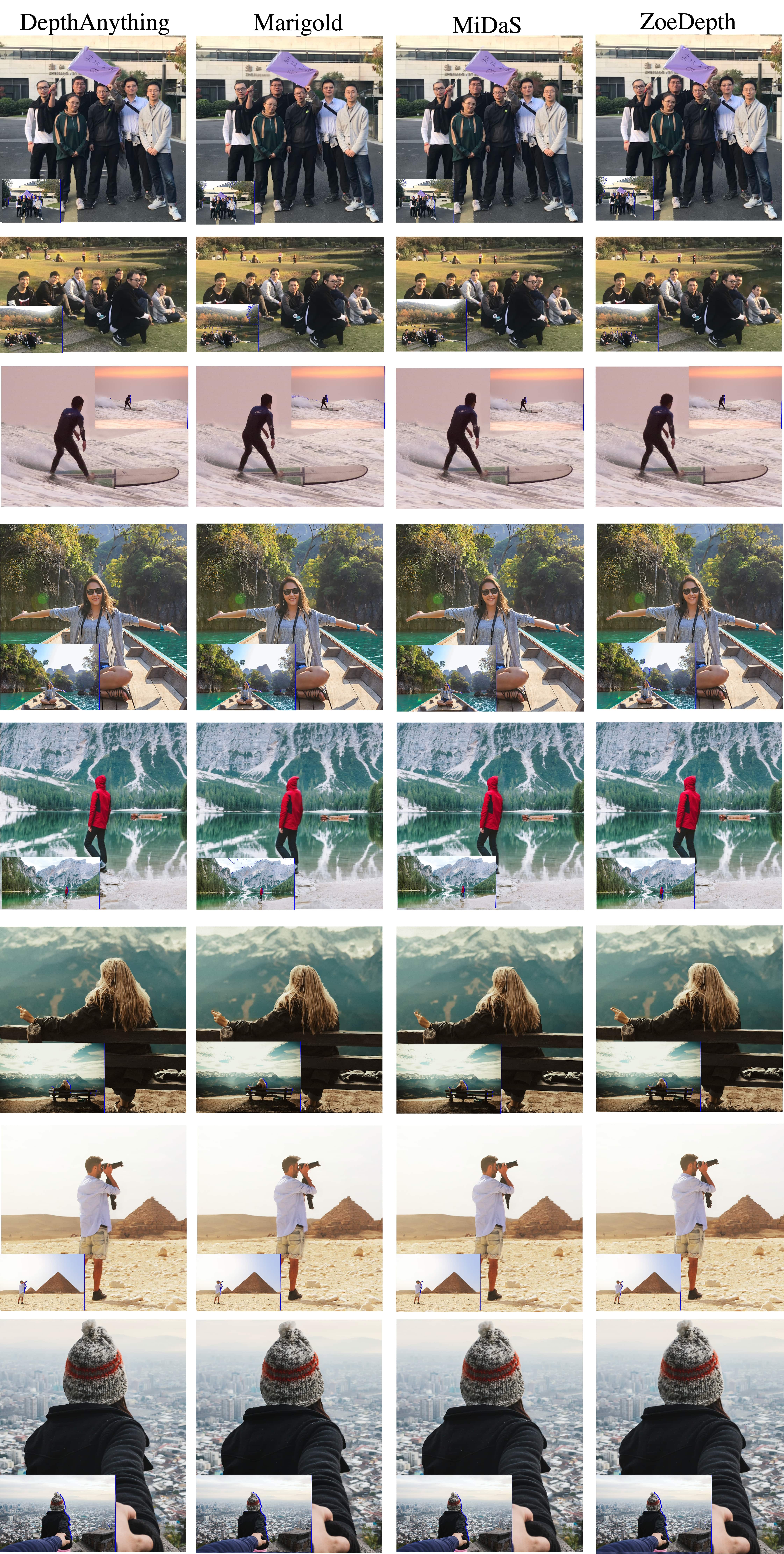}
    \caption{Visual comparison of different depth estimation methods. The zoomed image represents occluded regions.}
    \label{fig:depth}
\end{figure*}

\section{Video Synthesis}
\subsection{Computational Cost}
As shown in Table~\ref{tab:cost} with our 2.485 billion model.
\begin{table}[b]
\vspace{-1em}
\centering
\caption{Cost for a 1024x1024, 30-frame video on an A800 GPU.}
\label{tab:cost}
\resizebox{0.48\textwidth}{!}{
\large
\begin{tabular}{lccccccc}
\toprule
 & 3D-photography & Webui-depthmap &  P-NVS & CoPoNeRF & NVS-Solver & MVSplat &  Proposed \\
 \midrule
DVG & - & - & - & - & - &-& 51s \\
Inference & - & - & - & - &- & - & 1059s  \\
Total &10800s & 488s & 1503s & 519s & 833s & 41s &1110s \\
\bottomrule
\end{tabular}
}
\vspace{-1em}
\end{table}

\subsection{Quantitative Impact of Depth and Motion Estimation}
Table~\ref{tab:motionaccu} and Table~\ref{tab:depthaccu} show the impact of motion estimation methods and depth estimation methods, respectively.

\subsection{Quantitative comparison}
Table~\ref{tab:crafter} shows that our DVG/TIL module remains valid even with video depths. 

\begin{table}[b]
\centering
\caption{Quantitative comparison of motion estimation methods.}
\label{tab:motionaccu}
\resizebox{0.33\textwidth}{!}{
\large
\begin{tabular}{lccc}
\toprule
 & PWC-Net	 & RAFT(Proposed) & SEA-RAFT \\
 \midrule
FVD$\downarrow$ & 70.96&	67.09&	\textbf{66.49} \\
$E^{*}_{warp}$ $\downarrow$ & 3.661	& 3.374&	\textbf{3.302}  \\
\bottomrule
\end{tabular}
}
\vspace{-1em}
\end{table}

\begin{table}[t]
\centering
\caption{Quantitative comparison of depth estimation methods.}
\label{tab:depthaccu}
\resizebox{0.48\textwidth}{!}{
\large
\begin{tabular}{lccccc}
\toprule
 & DepthAnything	&Marigold&	ZoeDepth&DepthCrafter&MiDaS(Proposed) \\
 \midrule
FVD$\downarrow$ & 67.33 &69.12&	69.03&	\textbf{62.50}	&67.09 \\
$E^{*}_{warp}$ $\downarrow$ & 3.390&3.572&	3.365&	\textbf{3.333}	&3.374	  \\
\bottomrule
\end{tabular}
}
\vspace{-1em}
\end{table}


\begin{table}[t]
\centering
\caption{Ablation study of the DepthCrafter method}
\label{tab:crafter}
\resizebox{0.2\textwidth}{!}{
\begin{tabular}{lccc}
\toprule
TIL & DVG & FVD$\downarrow$ & $E^{*}_{warp} \downarrow$ \\
\midrule
 &   & 127.6 & 3.593 \\
\checkmark &   & 76.27 & 3.393 \\
 & \checkmark & 81.91 & 3.341 \\
\checkmark & \checkmark & 62.50 & 3.333 \\
\bottomrule
\end{tabular}
}
\end{table}



\subsection{Comparison with Other Methods}
\begin{figure*}[!ht]
    \vspace{-2cm}
    \centering
    \includegraphics[width=1.0\textwidth]{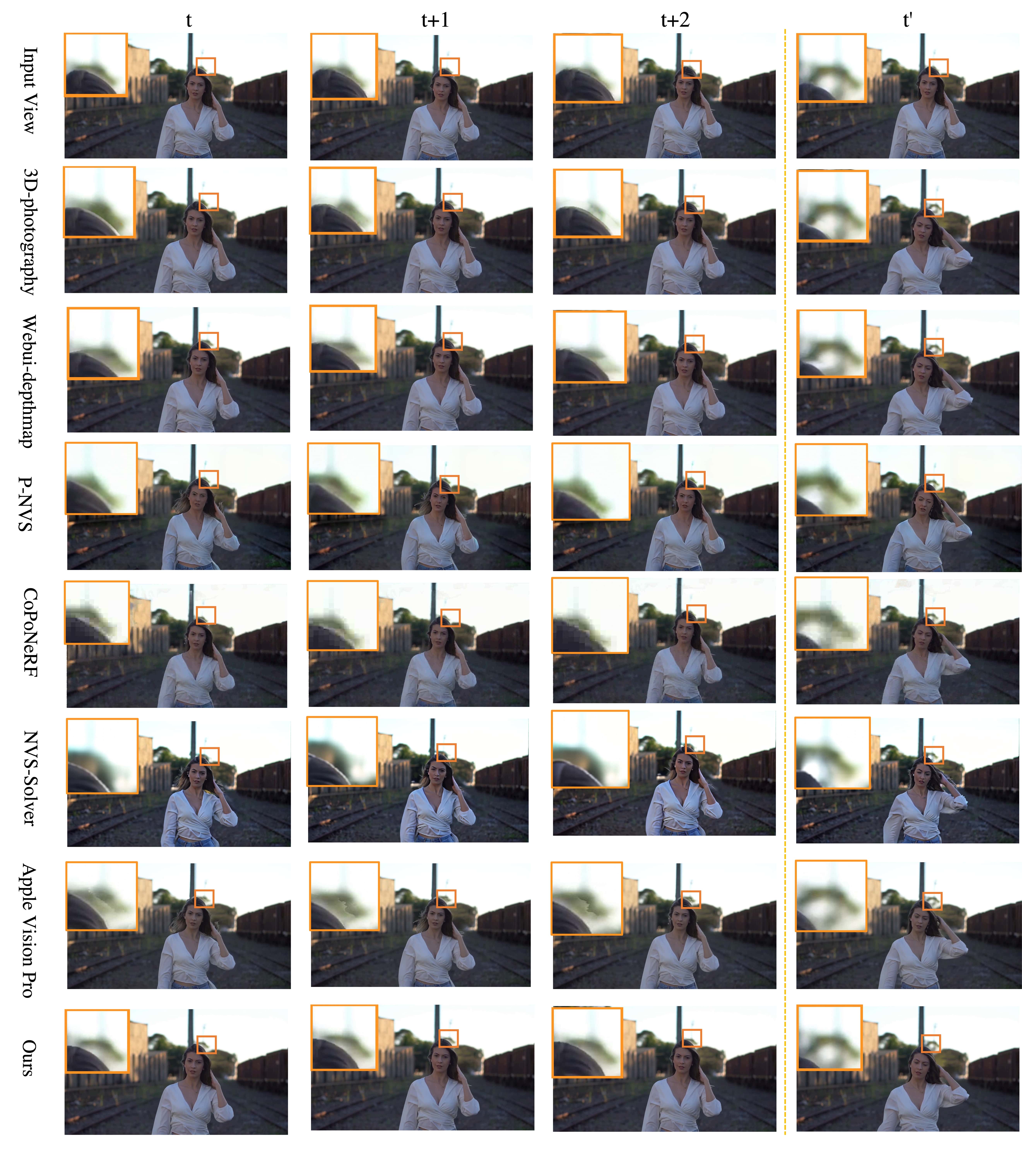}
    \centerline{(a)}
    \label{fig:video_com}
\end{figure*}
\begin{figure*}[!ht]
    \vspace{-2cm}
    \centering
    \includegraphics[width=1.0\textwidth]{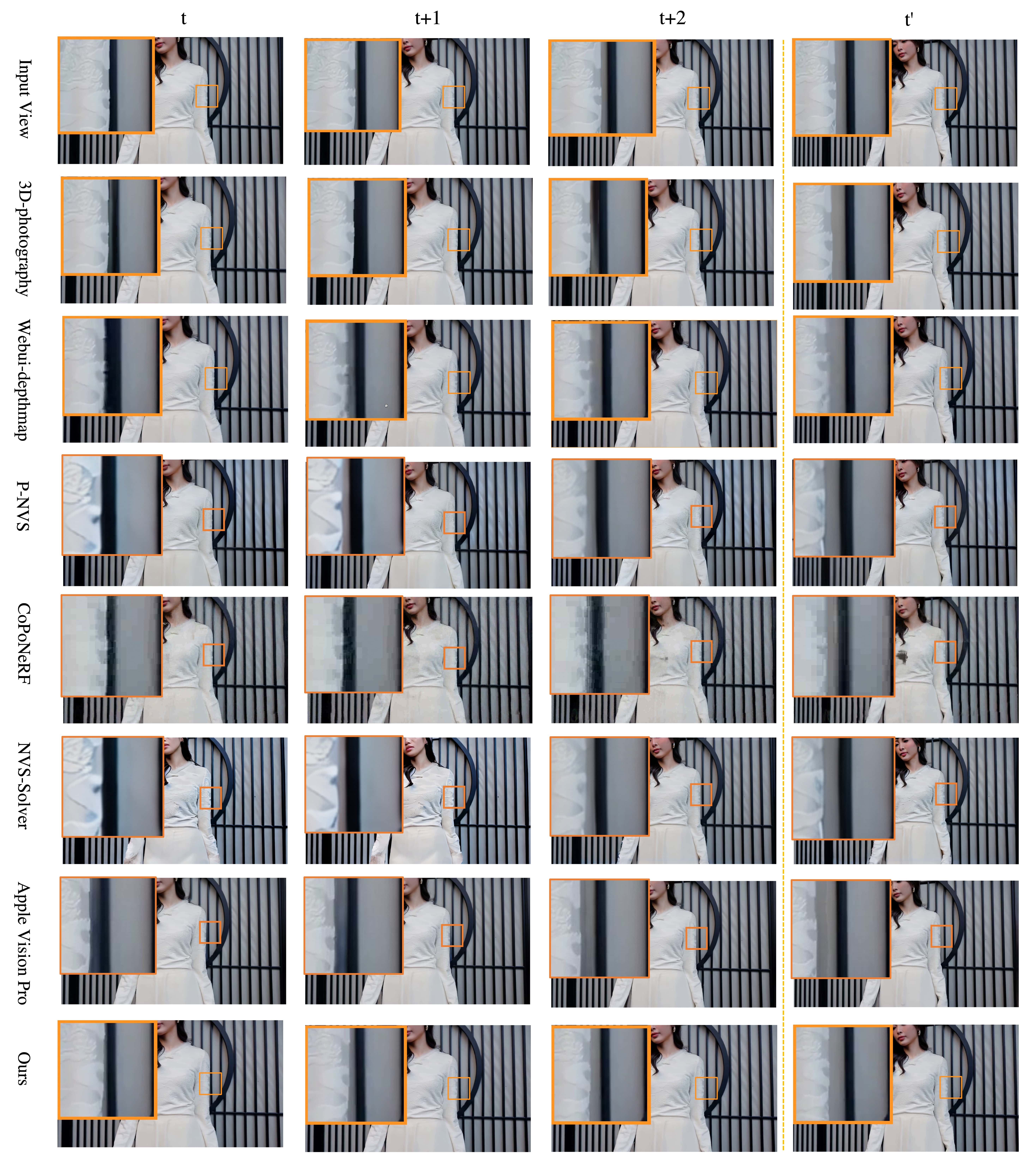}
    \centerline{(b)}
    \caption{Quantitative comparison with the other methods.The first three columns are adjacent frames with t, t+1, t+2, and the last column is the non-adjacent frame. The proposed method not only generates accurate and consistent content in the occluded regions among the adjacent frames, but also maintains consistency with the visible parts in the non-adjacent frame, as highlighted in the orange boxes.}
    \label{fig:video_com}
\end{figure*}

Video results compared with other methods are provided in folder ``$\bold{compare\_to\_others}$" and the corresponding visual comparison is shown in Figure~\ref{fig:video_com}.
The meaning of the file name is explained below:
\begin{itemize}
\item  input.mp4: input video
\item 3d-photo.mp4: video result using 3D-photography method
\item webui.mp4: video result of webui-depthmap method
\item P-NVS.mp4: video result of P-NVS method
\item coponerf.mp4: video result of CoPoNeRF method
\item nvs-solver.mp4: video result of NVS-Solver method
\item AVP.mp4: video result of Apple Vision Pro
\item ours.mp4: video result of our method
\end{itemize}

\subsection{Videos Under the Target Viewpoint}
The newly synthesized video results of our method are provided in folder ``$\bold{target\_view\_video}$".
\subsection{Stereo Video Results}
Furthermore, the final side-by-side video results are provided in folder ``$\bold{stereo\_video}$", which can be watched on VR device(AVP, Quest, Pico, etc.).

\label{sec:rationale}
%
%
\end{NoHyper}
\end{document}